\documentclass[wcp]{jmlr}

\usepackage{longtable}

\usepackage{booktabs}

\newcommand{\A}{\mathbf{A}}

\newcommand{\W}{\mathbf{W}}
\newcommand{\X}{\mathbf{X}}
\newcommand{\x}{\mathbf{x}}

\newcommand{\y}{\mathbf{y}}

\usepackage{pifont}%
\newcommand{\cmark}{\ding{51}}%
\newcommand{\xmark}{\ding{55}}%

\usepackage{amsmath, amssymb}

\usepackage{amsthm}
\usepackage{cleveref}
\usepackage{algorithm}
\usepackage{algpseudocode}
\usepackage{bm}
\usepackage{subfiles}

\makeatletter
\let\Ginclude@graphics\@org@Ginclude@graphics 
\makeatother

\jmlrvolume{189}
\jmlryear{2022}
\jmlrworkshop{ACML 2022}

\title[Short Title]{Multiple Imputation with Neural Network Gaussian Process for High-dimensional Incomplete Data}

  \author{\Name{Zongyu Dai} \Email{daizy@sas.upenn.edu} \\
  \Name{Zhiqi Bu} \Email{zbu@sas.upenn.edu}\\
  \addr Graduate Group in Applied Mathematics and Computational Science, University of Pennsylvania
  \AND
  \Name{Qi Long} \Email{qlong@upenn.edu}\\
  \addr Division of Biostatistics, University of Pennsylvania
 }

\editors{Emtiyaz Khan and Mehmet G\"{o}nen}

\begin{document}

\maketitle

\begin{abstract}
Missing data are ubiquitous in real world applications and, if not adequately handled, may lead to the loss of information and biased findings in downstream analysis. Particularly,  high-dimensional incomplete data with a moderate sample size, such as analysis of multi-omics data, present daunting challenges. Imputation is arguably the most popular method for handling missing data, though existing imputation methods have a number of limitations. Single imputation methods such as matrix completion methods do not adequately account for imputation uncertainty and hence would yield improper statistical inference. In contrast, multiple imputation (MI) methods allow for proper inference but existing methods do not perform well in high-dimensional settings. Our work aims to address these significant methodological gaps, leveraging recent advances in neural network Gaussian process (NNGP) from a Bayesian viewpoint. We propose two NNGP-based MI methods, namely MI-NNGP, that can apply multiple imputations for missing values from a joint (posterior predictive) distribution. The MI-NNGP methods are shown to significantly outperform existing state-of-the-art methods on synthetic and real datasets, in terms of imputation error, statistical inference, robustness to missing rates, and computation costs, under three missing data mechanisms, MCAR, MAR, and MNAR. Code is available in the
GitHub repository \href{https://github.com/bestadcarry/MI-NNGP}{https://github.com/bestadcarry/MI-NNGP}.  
\end{abstract}
\begin{keywords}
Missing Data, Multiple imputation, Neural Network Gaussian Processes, Statistical Inference \end{keywords}

\section{Introduction}
Missing data are frequently encountered and present significant analytical challenges in many research areas. Inadequate handling of missing data can lead to biased results in subsequent data analysis. For example, complete case analysis that uses only the subset of observations with all variables observed is known to yield biased results and/or loss of information as it does not utilize the information contained in incomplete cases \cite{little2019statistical}. Missing value imputation has become increasingly popular for handling incomplete data. Broadly  speaking, imputation methods can be categorized as single imputation (SI) or multiple imputation (MI). SI methods impute missing values for a single time which fails to adequately account for imputation uncertainty; in contrast, MI methods impute missing values multiple times by sampling from some (predictive) distribution to account for imputation uncertainty.  MI offers another significant advantage over SI in that it can conduct hypothesis testing or construct confidence intervals using multiply imputed datasets via Rubin's rule \cite{little2019statistical}. Of note, most popular imputation methods in the machine learning literature, such as matrix completion methods, are SI and hence expected to yield invalid statistical inference as shown in our numerical experiments.

When conducting imputation, it is important to know the mechanisms under which missing values originate, namely, missing completely at random (MCAR), missing at random (MAR), or missing not at random (MNAR) \cite{little2019statistical}. To be specific, MCAR means that the missingness does not depend on observed or missing data. While most imputation methods are expected to work reasonably well under MCAR, this assumption is typically too strong and unrealistic in practice, particularly for analysis of incomplete data from biomedical studies. MAR and MNAR are more plausible than MCAR. Under MAR, the missingness depends on only the observed values. Under MNAR, the missingness may depend on both observed and missing values, and it is well-known that additional structural assumptions need to be made in order to develop valid imputation methods under MNAR. 


Existing state-of-the-art imputation methods can be categorized into discriminative methods and generative methods. The former includes, but not limited to, MICE \cite{van2007multiple,deng2016multiple,zhao2016multiple}, MissForest \cite{stekhoven2012missforest}, KNN \cite{liao2014missing} and matrix completion \cite{mazumder2010spectral,hastie2015matrix}; the latter includes joint modeling \cite{schafer1997analysis,garcia2010pattern}, autoencoders \cite{ivanov2018variational,mattei2019miwae}, and generative adversarial networks \cite{9679967,yoon2018gain,lee2019collagan}. However, the existing imputation methods have several drawbacks. MICE imputes missing values through an iterative approach based on conditional distributions and requires to repeating the imputing procedures multiple times till convergence. MICE \cite{van2007multiple}, known to be computationally expensive, tends to yield poor performance and may become computationally infeasible for high-dimensional data with high missing rates. Joint modeling (JM), another classical imputation method, relies on strong assumptions for the data distribution. Its performance also deteriorates rapidly as the feature dimension increases. SoftImpute \cite{mazumder2010spectral}, a matrix completion method,  conducts single imputation based on the low-rank assumption, leading to underestimating the uncertainty of imputed values. In recent years, many deep learning-based imputation methods have been proposed. As the most representative one, GAIN \cite{yoon2018gain} can handle mixed data types. However, the appliance for GAIN in practice is limited as it is valid only under MCAR. Most recently, importance-weighted autoencoder based method MIWAE \cite{mattei2019miwae} and not-MIWAE \cite{ipsen2020not} can deal with MAR and MNAR mechanism, respectively. Plus, optimal transport-based methods \cite{muzellec2020missing} including Sinkhorn and Linear RR have been shown to outperform other state-of-the-art imputation methods under MCAR, MAR and MNAR. However, above methods are shown to exhibit appreciable bias in our high-dimensional data experiments. Moreover, Linear RR imputes missing values iteratively like MICE, hence is inherently not suitable for high-dimensional incomplete data.

To address the limitations of existing imputation methods, we leverage recent developments in neural network Gaussian process (NNGP) theory \cite{williams1997computing,lee2017deep,novak2018bayesian,neuraltangents2020} to develop a new robust multiple imputation approach. 
The NNGP theory can provide explicit posterior predictive distribution for missing values and allow for Bayesian inference without actually training the neural networks, leading to substantial savings in training time. Here we take $L$-layer fully connected neural networks as an example. Suppose a neural network has layer width $n^l$(for hidden layer $l$), activation function $\phi$ and centered normally distributed weights and biases with variance $\frac{\sigma_w^2}{n^l}$ and $\sigma_b^2$ at layer $l$. When each hidden layer width goes to infinity, each output neuron is a Gaussian process $\mathcal{GP}(0,\mathcal{K}^L)$ where $\mathcal{K}^L$ is deterministically computed from $L, \phi, \sigma_w, \sigma_b$. Details can be found in appendix.

{\bf Our contribution:} Our proposed deep learning imputation method, Multiple Imputation through Neural Network Gaussian Process (MI-NNGP), is designed for the high-dimensional data setting in which the number of variables/features can be large whereas the sample size is moderate. This setting is particularly relevant to analysis of incomplete high-dimensional -omics data in biomedical research where the number of subjects is typically not large. MI-NNGP is the first deep learning based method that yields satisfactory performance in statistical inference for high-dimensional incomplete data under MAR. Empirically speaking, MI-NNGP demonstrates strong performance on imputation error, statistical inference, computational speed, scalability to high dimensional data, and robustness to high missing rate. \ref{table: performance} summarizes the performance of MI-NNGP in comparison with several existing state-of-the-art imputation methods. 

\begin{table}[!htb]
	\centering
	\begin{tabular}{|c|c|c|c|c|c|c|c|}
	\hline 
	\text{Models}   &\text{MI} & \text{Imp Error} & \text{Inference} & Scalability \\
	\hline
		MI-NNGP  &\cmark  &\cmark &\cmark &\cmark  \\

	\hline 
    
    Sinkhorn  &\cmark  &\textbf{?} &\textbf{?} &\xmark   \\
    \hline
    Linear RR  &\cmark  &\textbf{?} &\textbf{?} &\xmark   \\
    	\hline 
	MICE  &\cmark  &\cmark &\textbf{?}  &\xmark   \\
		\hline 
	SoftImpute   &\xmark  &\cmark &\xmark &\textbf{?} 
 \\
 \hline 
 	MIWAE   &\cmark &\cmark &\textbf{?}  &\textbf{?} 

 \\
		\hline 
	GAIN  &\cmark  &\xmark &\xmark &\xmark \\

	\hline 
	\end{tabular}
\caption{Summary of imputation methods. Imp Error refers to imputation error. Question mark indicates that the performance depends on specific settings.}
\label{table: performance}
\vspace{-0.2in}
\end{table}

\vspace{-0.1in}
\section{Problem Setup}
\label{section:problem setup}

\begin{figure}[b]
  \centering
  \includegraphics[width=0.7\linewidth,height=3.5cm]{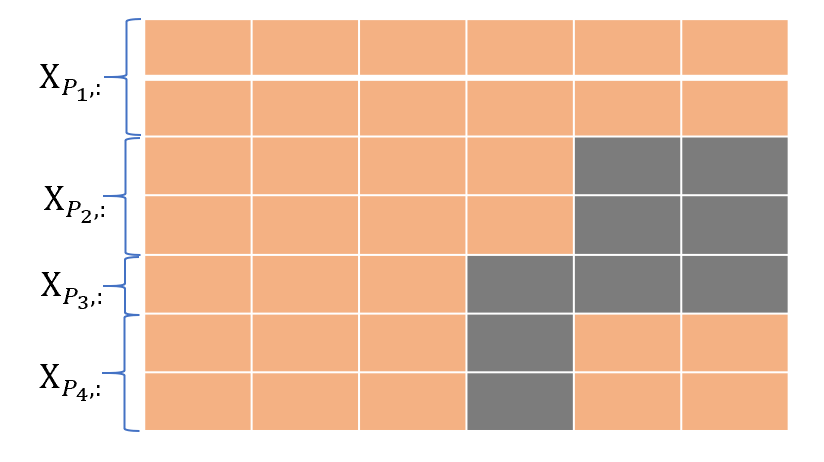}
  \caption{Multivariate 4-pattern missing data. Orange squares represent observed data and gray squares represent missing data.}
 \label{table:4 pattern}
\end{figure}

To fix ideas, we consider the multivariate $K$-pattern missing data, meaning that observations can be categorized into $K$ patterns according to which features have missing values. Within each pattern, a feature is either observed in all cases or missing in all cases as visualized in \Cref{table:4 pattern} which provides an illustration of $4$-pattern missing data. As a motivation, multivariate $K$-pattern missing data are often encountered in medical research. For example, the Alzheimer’s Disease Neuroimaging Initiative (ADNI) collected high-dimensional multi-omics data and each -omics modality is measured in only a subset of total cases, leading to the multivariate $K$-pattern missing data. In addition, a general missing data pattern, after some rearranging of rows, can be converted to $K$-pattern missing data. 

Suppose we have a random sample of $n$ observations with $p$ variables. Denote the $n \times p$ observed data matrix by $\X$, which may include continuous and discrete values and where $\X_{i,j}$ is the value of $j$-th variable/feature for $i$-th case/observation. Let $\X_{i,:}$ and $\X_{:,j}$ denote the $i$-th row vector and $j$-th column vector, respectively. Since some elements of $\X$ are missing, the $n$ observations/cases can be grouped into $K$ patterns (i.e. $K$ submatrices) $\X_{P_k,:}$ for $k\in[K]$; see the illustrative example in \Cref{table:4 pattern}. Here $P_k$ is the index set for the rows in $\X$ which belong to the $k$-th pattern. Without loss of generality, we let $\X_{P_1,:}$ denote the set of complete cases for which all features are observed. We define  $\X_{P_{-k},:}=\X\backslash\X_{P_k,:}$ as the complement data matrix for $\X_{P_k,:}$. We denote by $\textbf{obs}(k)$ and $\textbf{mis}(k)$ the index sets for the columns in $\X_{P_k,:}$ that are fully observed and fully missing, respectively. 

\section{Multiple Imputation via Neural Network Gaussian Process}

In this section, we develop two novel MI methods for multivariate $K$-pattern missing data based on NNGP. Specifically, we first propose MI-NNGP1 which imputes each missing pattern by  exploiting information contained in the set of complete cases ($\X_{P_1,:}$). We then propose MI-NNGP2, which imputes each missing pattern iteratively by utilizing the information contained in all observed data. We further improve both methods by incorporating a bootstrap step. 

\subsection{Imputing Missing Data from an Alternative Viewpoint}
\label{section: new idea}
MICE is a quite flexible MI method as it learns a separate imputation model for each variable, in an one by one manner. However, MICE is extremely slow in high dimensional setting and incapable of learning the features jointly (hence underestimate the interactions). To overcome these drawbacks, we leverage the NNGP to efficiently impute all missing features of one observation simultaneously. To this end, we propose a `transpose' trick when using NNGP. We regard each column/feature of $X$, instead of each row/case of $X$, as a `sample', so that we draw all features jointly instead of drawing all cases jointly as in the conventional NNGP. As demonstrated in our experiments, this appealing property makes our MI-NNGP methods scalable to high-dimensional data where $p$ can be very large.

As a building block, we first consider imputing the $k$-th pattern of missing data ($k=1,\ldots,K$). We define a training set $\left\{\left(\X_{\text{IS},t},\X_{P_k,t}\right)\right\}_{t \in \text{obs}(k)}$ and a test set $\left\{\left(\X_{\text{IS},t},\X_{P_k,t} \right)\right\}_{t \in \text{mis}(k)}$, where $\X_{\text{IS},t}$ is the input data point and $\X_{P_k,t}$ is the output target. Here the index set `IS' represents the cases included as input, which depends on the specific algorithm used. For example, MI-NNGP1 uses $P_1$ as the IS set; see details in \cref{section: minngp1}. Of note, our goal is to predict the test set label $\left\{\X_{P_k,t}\right\}_{t \in \text{mis}(k)}$ which is missing. Denote the size of the training and test set by $|\text{obs}(k)|$ and $|\text{mis}(k)|$,  respectively. 

Given the training and the test sets, we specify a neural network $f_k:\mathbb{R}^{|\text{IS}|}\to \mathbb{R}^{|P_k|}$ for the $k$-th pattern. Therefore, each case in the $k$-th pattern (say the $i$-th case, $i\in P_k$) corresponds to an output neuron (say the $j$-th output component): $f_k^j\left(\X_{\text{IS},t}\right) = \X_{i,t}$ for $t\in [p]$, if we assume all observed values are noise-free. By considering infinitely-wide layers, each output component/neuron can be regarded as an independent Gaussian process $\mathcal{GP}(0,\mathcal{K})$ in terms of its input. Here the covariance function $\mathcal{K}$ is uniquely determined by two factors, the neural network architecture (including the activation function) and the initialization variance of weight and bias \cite{lee2017deep}. Hence, for the $j$-th output component of the network, $f_k^j$, we know 
\begin{align}  
\begin{bmatrix}
f_k^j\left(\X_{\text{IS},1}\right) \\
\vdots  \\
f_k^j\left(\X_{\text{IS},p}\right)
\end{bmatrix} \sim \mathcal{N}\left(0,\Sigma \right)
\end{align}
where $\Sigma\in\mathbb{R}^{p\times p}$ whose $(u,v)$-th element is $\mathcal{K}\left(\X_{\text{IS},u},\X_{\text{IS},v}\right)$. Hence, we get: 
\begin{align} \label{equation: gaussian}  
\begin{bmatrix}
\X_{i,\text{obs}(k)}^{\top} \\
\X_{i,\text{mis}(k)}^{\top}
\end{bmatrix} \sim \mathcal{N}\left(0,\begin{bmatrix}
\Sigma_{11} & \Sigma_{12} \\
\Sigma_{21} & \Sigma_{22} 
\end{bmatrix}\right)
\end{align}
where the block structure corresponds to the division between the training and the test sets. Specifically, $\Sigma_{11}=\mathcal{K}\left(\X_{\text{IS},\text{obs}(k)},\X_{\text{IS},\text{obs}(k)}\right)$,  $\Sigma_{22}=\mathcal{K}\left(\X_{\text{IS},\text{mis}(k)},\X_{\text{IS},\text{mis}(k)}\right)$, $\Sigma_{12}=\Sigma_{21}^{\top} = \mathcal{K}\left(\X_{\text{IS},\text{obs}(k)},\X_{\text{IS},\text{mis}(k)}\right)$, where $\Sigma_{11}\in\mathbb{R}^{|\text{obs}(k)|\times|\text{obs}(k)|}$ is composed of $\mathcal{K}\left(\X_{\text{IS},u},\X_{\text{IS},v}\right)$ for $u,v \in \text{obs}(k)$. Then, \eqref{equation: gaussian} indicates that the missing values $\X_{i,\text{mis}(k)}$, conditioned on the known values (either observed or previously imputed), follow a joint posterior distribution,
\begin{align} \label{equation: posterior}
\begin{split}
    \X_{i,\text{mis}(k)}^{\top}  \Big| \X_{i,\text{obs}(k)}^{\top},\X_{\text{IS},\text{obs}(k)},\X_{\text{IS},\text{mis}(k)} \sim \\ \mathcal{N}\left(\Sigma_{21}\Sigma_{11}^{-1} \X_{i,\text{obs}(k)}^{\top},\Sigma_{22}-\Sigma_{21}\Sigma_{11}^{-1}\Sigma_{12} \right) 
\end{split}    
\end{align}
Equation \eqref{equation: posterior} allows us to multiply impute $\X_{i,\text{mis}(k)}$. Here we emphasize that NNGP is not a linear method. The imputed values are drawn from a Gaussian distribution. 

Note that inverting $\Sigma_{11}$ is a common computational challenge. The time complexity is cubic in $p$. We use the efficient implementation in \textbf{neural tangents} \cite{neuraltangents2020} to solve this problem. For the setting in our paper, when $p\sim 10000$, inverting $\Sigma_{11}$ just cost several seconds with GPU P100 and 16G memory.

\subsection{MI-NNGP1 --- Direct imputation}

\label{section: minngp1}
\begin{figure}
  \centering
  \includegraphics[width=7.5cm,height=6cm]{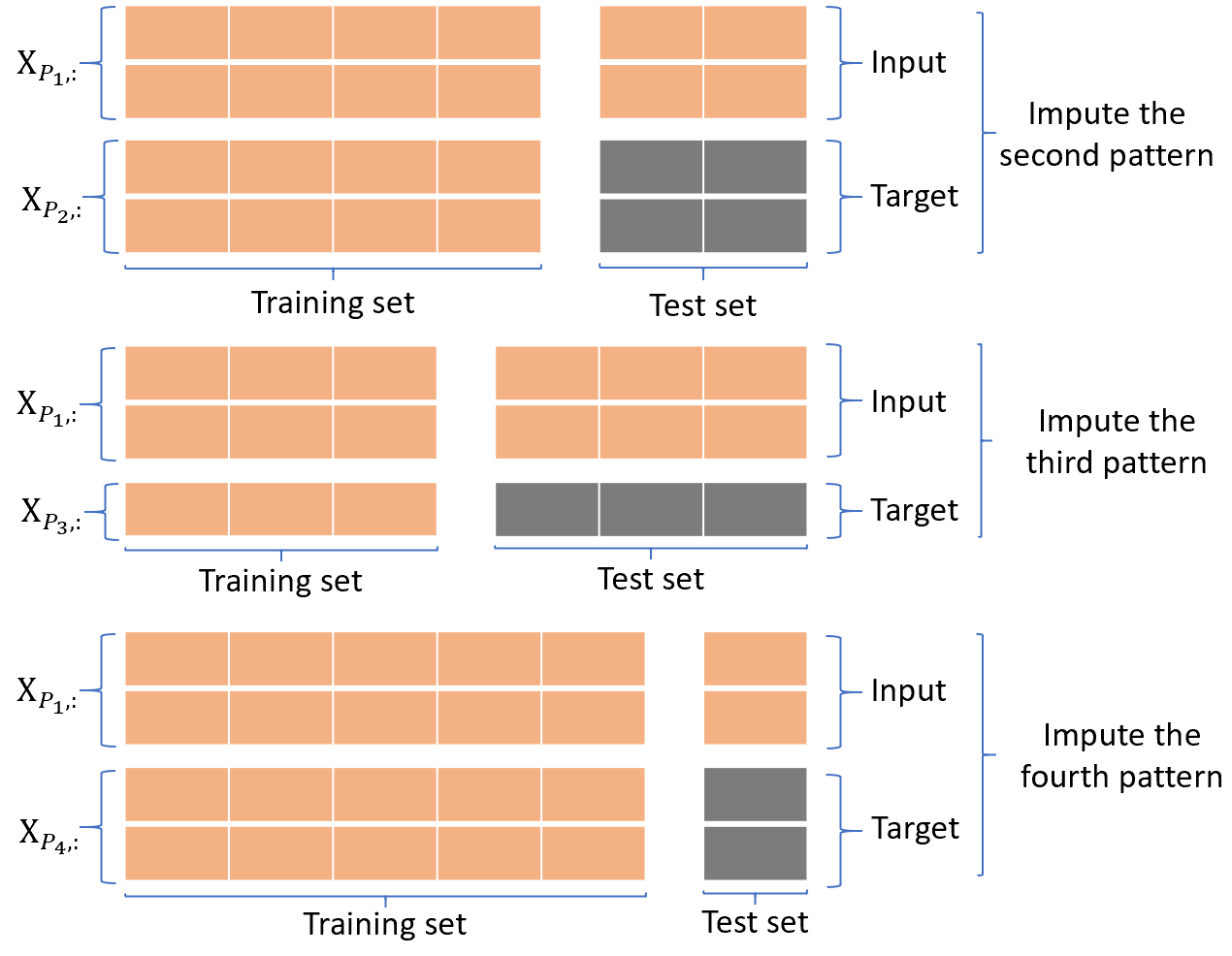}
  \caption{MI-NNGP1 applied to the four-pattern missing data in \Cref{table:4 pattern}}
  \label{figure: migp1}
\end{figure}

Our first algorithm is MI-NNGP1 that uses only the complete cases to impute all missing values. More precisely, to impute the missing values in the $k$-th pattern, we select $P_1$ as our IS set. Hence, for each $k$, we essentially divide all features in the first and the $k$-th pattern into the training set $\left\{\left(\X_{P_1,t},\X_{P_k,t}\right)\right\}_{t \in \text{obs}(k)}$ and the test set $\left\{\left(\X_{P_1,t},\X_{P_k,t} \right)\right\}_{t \in \text{mis}(k)}$ as (input,target) pairs. Following the steps described in \Cref{section: new idea}, the covariance matrices are
\begin{align*}
    \Sigma_{11}&=\mathcal{K}(\X_{P_1,\text{obs}(k)},\X_{P_1,\text{obs}(k)}),\\ \Sigma_{22}&=\mathcal{K}(\X_{P_1,\text{mis}(k)},\X_{P_1,\text{mis}(k)}), \\ \Sigma_{12}&=\Sigma_{21}^\top = \mathcal{K}(\X_{P_1,\text{obs}(k)},\X_{P_1,\text{mis}(k)}).
\end{align*}
We then draw the imputed values multiple times from the posterior distribution in \eqref{equation: posterior} for all $i \in P_k$. The whole process is summarized in \Cref{MIGP1} and \Cref{figure: migp1}. We use the same neural network architecture and the initialization variance of weight and bias for imputing each pattern. Note that the kernel function $\mathcal{K}$ does not depends on the length of the input or that of the output, so the same $\mathcal{K}$ is shared across all patterns. The time complexity of MI-NNGP1 is $Kp^3$ for imputing a K-pattern missing data.  

\subsection{MI-NNGP2 --- Iterative imputation}
In contrast to MI-NNGP1, which imputes each incomplete case basing only on the complete cases, we here propose MI-NNGP2 to impute through an iterative approach that leverages the information contained in incomplete cases. As such, MI-NNGP2  works with a small number of complete cases or even when there is no complete case.

MI-NNGP2 requires an initial imputation $\widehat{\X}$ for the entire data. This can be done by either MI-NNGP1 (if complete cases exist), column mean imputation, or another imputation method. Starting from the initial imputation $\widehat{\X}$, MI-NNGP2 imputes the missing part of each pattern and updates $\widehat{\X}$ iteratively: e.g. the imputed values of $k$-th pattern is used to impute the missing values of the $(k+1)$-th pattern. 
To be more precise, when imputing the $k$-th pattern, we select $P_{-k}$ as the IS set. Hence, we have the training set $\left\{\left(\widehat{\X}_{P_{-k},t},\widehat{\X}_{P_k,t}\right)\right\}_{t \in \text{obs}(k)}$ and test set $\left\{\left(\widehat{\X}_{P_{-k},t},\widehat{\X}_{P_k,t}\right)\right\}_{t \in \text{mis}(k)}$ of (input,target) pairs. Then we calculate the covariance matrix
\begin{align*}
  \Sigma_{11}&=\mathcal{K}(\widehat{\X}_{P_{-k},\text{obs}(k)},\widehat{\X}_{P_{-k},\text{obs}(k)}),\\ \Sigma_{22}&=\mathcal{K}(\widehat{\X}_{P_{-k},\text{mis}(k)},\widehat{\X}_{P_{-k},\text{mis}(k)}),\\ \Sigma_{12}&=\Sigma_{21}^\top = \mathcal{K}(\widehat{\X}_{P_{-k},\text{obs}(k)},\widehat{\X}_{P_{-k},\text{mis}(k)})  
\end{align*} 
and impute the $k$-th pattern in $\widehat{\X}$ by drawing $\widehat{\X}_{i,\text{mis}(k)}$ form the posterior distribution \eqref{equation: posterior} for each $i\in P_k$. This method is described by \Cref{MIGP2}. Similar to MI-NNGP1, $\mathcal{K}$ is shared across all patterns in MI-NNGP2. To conduct multiple imputation, we do not record the imputed values in the first $N$ cycles. After this burn-in period, we choose $\widehat{\X}$ at every $T$-th iteration.

\begin{algorithm}[!htb]
\caption{MI-NNGP1: direct imputation}\label{MIGP1}
  \textbf{Input:} Incomplete matrix $\X$ , imputation times $M$, neural network architecture, initialization variance of weight and bias\\
 \textbf{Output:} $M$ Imputed matrix
\begin{algorithmic}[1]
\State  Calculate the corresponding kernel function $\mathcal{K}$ of the network

\State Create $M$ copies of $\X$, denoted as $\widehat{\X}^m$, $m\in M$.

\For{\texttt{$k \in \{2,\dots,K\}$}}

\State Calculate the matrix $\Sigma_{11},\Sigma_{21},\Sigma_{12}$ and $\Sigma_{22}$. 


\For{\texttt{$i \in P_k$}} 

\State Draw $\widehat{\X}_{i,\text{mis}(k)}$ for $M$ times from $\mathcal{N}\left(\Sigma_{21}\Sigma_{11}^{-1}  \X_{i,\text{obs}(k)}^{\top},\Sigma_{22}-\Sigma_{21}\Sigma_{11}^{-1}\Sigma_{12} \right)$ 

\State Update $\widehat{\X}^m$ with one $\widehat{\X}_{i,\text{mis}(k)}$ for each $m\in [M]$.

\EndFor

\EndFor
\State Output $\widehat{\X}^m$, $m\in [M]$.
\end{algorithmic}
\end{algorithm}

The time complexity of MI-NNGP2 is $(N+MT)Kp^3$ where $M$ represents imputation times and usually selected as 10. Here $N, M, T$ are bounded by constant and much smaller than $K$ and $p$. In our experiment, $N=2$ and $T=1$ leads to excellent performance. It is important to note that although MI-NNGP2 imputes missing values iteratively, the time cost is expected to increase only modestly compared to MI-NNGP1.

\begin{algorithm}[!htb]
\caption{MI-NNGP2: iterative imputation}\label{MIGP2}

\textbf{Input:} Initial imputation $\widehat{\X}$, imputation times $M$, burn-in period $N$, sampling interval $T$, neural network architecture, initialization variance of weight and bias \\
\textbf{Output:} $M$ Imputed matrix

\begin{algorithmic}[1]

\State Calculate the corresponding kernel function $\mathcal{K}$ of the network.

\For{\texttt{$l \in \{1,\dots,N+MT\}$}}

\For{\texttt{$k \in \{2,\dots,K\}$}}
\State Calculate the matrix $\Sigma_{11},\Sigma_{21},\Sigma_{12}$ and $\Sigma_{22}$.

\For{\texttt{$i \in P_k$}} 

\State Draw $\widehat{\X}_{i,\text{mis}(k)}$ from $\mathcal{N}\left(\Sigma_{21}\Sigma_{11}^{-1}  \widehat{\X}_{i,\text{obs}(k)}^{\top},\Sigma_{22}-\Sigma_{21}\Sigma_{11}^{-1}\Sigma_{12} \right)$ 

\State Update $\widehat{\X}$ with $\widehat{\X}_{i,\text{mis}(k)}$ 

\EndFor

\EndFor

\If{$l>N$ and $T\big|(l-N)$}
 \State output $\widehat{\X}$ 
 \EndIf
\EndFor
\end{algorithmic}
\end{algorithm}

\subsection{MI-NNGP with bootstrapping}

\begin{figure*}[tb]
  \centering
  \includegraphics[width=12cm]{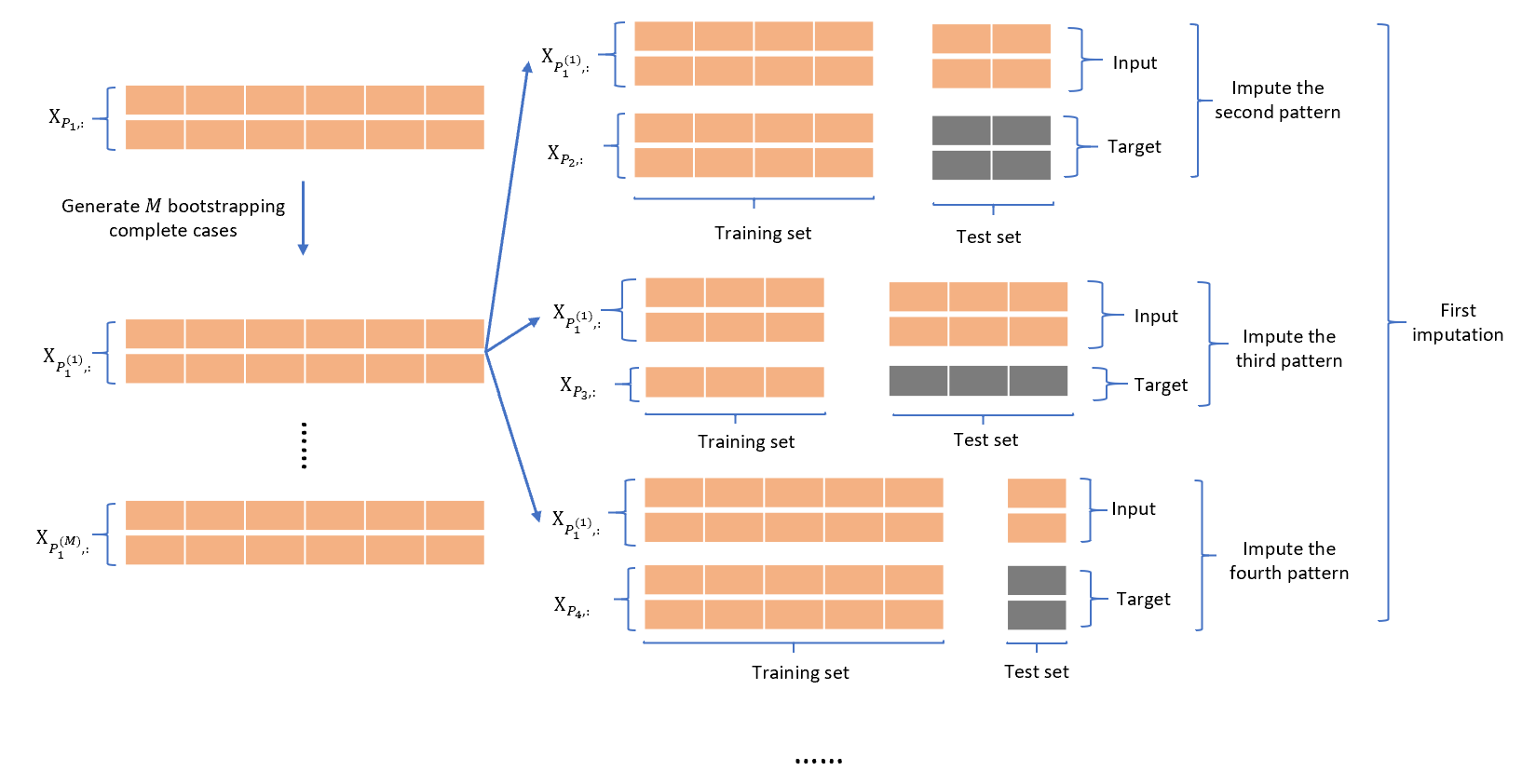}
  \caption{MI-NNGP1 with bootstrapping applied to the four-pattern missing data in \Cref{table:4 pattern}}
  \label{figure: minngp1-BS}
\end{figure*}

In the missing data literature, a bootstrap step has been incorporated in nonparametric imputation methods to better account for imputation uncertainty and improve statistical inference. The MI-NNGP methods can also be enhanced by including a bootstrap step. We illustrate this idea for MI-NNGP1. For each incomplete case, MI-NNGP1 essentially draws multiple imputations from the same posterior distribution. However, this may underestimate the uncertainty of imputed values. To overcome this potential drawback, we construct bootstrapping sets of $P_1$, denoted as $P_1^{(m)}$ for $m\in[M]$. Each bootstrapping set $P_1^{(m)}$ serves as the IS set for the $m$-th imputation as visualized in \Cref{figure: minngp1-BS}. We remark that using the bootstrapping adds negligible additional cost but usually improves the statistical coverage. Similarly, a bootstrap step can also be combined with MI-NNGP2. We can first use MI-NNGP1 with bootstrapping to generate multiple initial imputations and then run MI-NNGP2 multiple times from these initial imputations, where we choose $M=1$ in each track of MI-NNGP2.     

\section{Experiments}
We evaluate the performance of the MI-NNGP methods through extensive synthetic and real data experiments. The details about the experiment setup are provided in Appendix B and C. A brief outline of the synthetic data experiments is as follows. In each synthetic data experiment, we generate the data matrix from a pre-specified data model and then generate missing values under MCAR, MAR or MNAR. We apply an imputation method to each incomplete dataset; SI methods yield one imputed dataset and MI methods yield multiple imputed datasets. To assess the statistical inference performance, each imputed dataset is used to fit a regression model to obtain regression coefficient estimates and Rubin's rule \cite{little2019statistical} is used to obtain the final regression coefficient estimates $\hat{\bm\beta}$, their standard errors SE($\hat{\bm\beta}$), and $95\%$ confidence intervals.

\subsection{Imputation Methods Compared}
\label{section: imputation methods compared}

\textbf{Benchmarks.} (i) \textbf{Complete data} analysis assumes there is no missingness and directly fit a regression on the whole dataset. (ii) \textbf{Complete case} analysis does not incorporate imputation and fit a regression using only the complete cases. (iii) Column mean imputation (\textbf{ColMean Imp}) is feature-wise mean imputation. Here the complete data analysis serves as a golden standard, representing the best result an imputation method can possibly achieve. The complete case analysis and column mean imputation, two naive methods, are used to benchmark potential bias and loss of information (as represented by larger SE/SD) under MAR and MNAR.   

\textbf{State-of-the-art.} (iv) \textbf{MICE} (multiple imputation through chained equations) \cite{van2007multiple} is an popular and flexible multiple imputation method and has good empirical results and requires little tuning, but it fails to scale to high dimensional settings.  (v) \textbf{GAIN} \cite{yoon2018gain} is a generative neural network (GAN)\cite{goodfellow2014generative} based imputation method. (vi) \textbf{SoftImpute} \cite{mazumder2010spectral} is a matrix completion method and uses iterative soft-thresholded SVD to conduct missing data imputation. (vii) \textbf{Sinkhorn} \cite{muzellec2020missing} is a direct non-parametric imputation method which leverages optimal transport distance. (viii)  \textbf{Linear RR} \cite{muzellec2020missing} is a Round-Robin Sinkhorn Imputation. Similar to MICE, Linear RR iteratively impute missing features using other features in a cyclical manner. (ix) \textbf{MIWAE} \cite{mattei2019miwae} is a importance weighted autoencoder \cite{burda2015importance} (IWAE) based imputation method. 

\textbf{Our methods.} (x) \textbf{MI-NNGP1} uses the complete cases to conduct direct imputation as detailed in \Cref{MIGP1}. (xi) \textbf{MI-NNGP2} corresponds to \Cref{MIGP2} with burn-in period $N=10$ and sampling interval $T=1$. (xii)  \textbf{MI-NNGP1-BS} is MI-NNGP1 with an added bootstrap step.  (xiii) \textbf{MI-NNGP2-BS} runs MI-NNGP2 for multiple times with different initial imputations from MI-NNGP1-BS. In each parallel run of MI-NNGP2, we choose $N=2$ and $M=1$.

\subsection{Performance Metrics}

All performance metrics are averaged over 100 Monte Carlo (MC) datasets or repeats unless noted otherwise. To evaluate imputation accuracy and computational costs, we report the imputation mean squared error (Imp MSE) and the computing time in seconds per imputation (Time(s)). To evaluate statistical inference performance, we report bias of $\hat\beta_1$ denoted by Bias($\hat\beta_1$), standard error of $\hat\beta_1$ denoted by SE($\hat\beta_1$),  and coverage rate of the $95\%$ confidence interval for $\hat\beta_1$ denoted by CR($\hat\beta_1$), where $\hat\beta_1$ is one of the regression coefficients in the regression model fitted using imputed datasets. Some remarks are in order. A CR($\hat\beta_1$) that is well below the nominal level of $95\%$ would lead to inflated false positives, an important factor contributing to lack of reproducibility in research. To benchmark SE($\hat\beta_1$), we also report the  standard deviation of  $\hat\beta_1$ across 100 MC datasets denoted by SD($\hat\beta_1$), noting that a well-behaved SE($\hat\beta_1$) should be close to SD($\hat\beta_1$). In addition, while  we know the true value of $\beta_1$ and can report its bias in the synthetic data experiments, we do not know the true value of $\beta_1$ and cannot report its bias in the real data experiment.

\subsection{Synthetic data} \label{section: simulation}

The synthetic data experiments are conducted for low and high data dimensions, varying missing rates, and continuous and discrete data. In this section, we summarize the results from high dimensional settings (i.e. $p>n$) under MAR. Additional simulation results are included in appendix.

\begin{table*}[!htb]
	\centering
	\scalebox{0.8}{
	\begin{tabular}{|c|c|c|c|c|c|c|c|c}
	\hline 
	\text{Models} &\text{Style} &\text{Time(s)} & \text{Imp MSE} & \text{Bias}($\hat\beta_1$) & \text{CR($\hat\beta_1$}) & SE($\hat\beta_1$) & SD($\hat\beta_1$) \\ 
	\hline 
	
	SoftImpute &SI &15.1 &0.0200 &-0.0913 &0.78  &0.1195 &0.1624 \\

    GAIN &SI &39.0 &0.8685 & 0.6257 & 0.18 &0.1463 &0.5424 \\
    
    MIWAE &SI &46.3 &0.0502 & 0.0731 & 0.90 &0.1306 &0.1379 \\
    
    
    Linear RR & SI &3134.7 &0.0661 & 0.1486 & 1.00 &0.1782 &0.1011 \\
    
	MICE & MI & 37.6 & 0.0234 & \textbf{-0.0061} &\textbf{0.93}  &0.1167 &0.1213 \\
    	
    Sinkhorn &MI &31.2 &0.0757 &\textbf{0.0205} &\textbf{0.96} &0.1864 &0.1636 \\
    
	MI-NNGP1 &MI &4.9 &\textbf{0.0116} &\textbf{0.0077} &\textbf{0.92} &0.1147 &0.1223 \\
	
	MI-NNGP1-BS &MI &3.4 &\textbf{0.0149} &\textbf{0.0156} &\textbf{0.96} &0.1297 &0.1182 \\
    
    MI-NNGP2 &MI &5.7 &\textbf{0.0086} &\textbf{0.0012} &\textbf{0.96} &0.1179 &0.1170  \\
    
    MI-NNGP2-BS &MI &13.9 &\textbf{0.0094} &\textbf{0.0010} &\textbf{0.95} &0.1173 &0.1206  \\
    
	\hline 
    Complete data &-&-&-&-0.0027 & 0.90 &0.1098 &0.1141\\
	Complete case &-&-&-&0.2481 & 0.88 &0.3400 & 0.3309 \\
	ColMean Imp &SI&-& 0.1414 & 0.3498 &0.72 &0.2212 &  0.1725\\
	\hline 
	\end{tabular}}
\caption{Gaussian data with $n=200$ and $p=251$ under MAR. Approximately $\textbf{40\%}$ features and $\textbf{90\%}$ cases contain missing values. Detailed simulation setup information is in appendix.}
\label{p250 Gaussian MAR}
\end{table*}

\begin{table*}[!htb]
	\centering
	\scalebox{0.7}{
	\begin{tabular}{|c|c|c|c|c|c|c|c|c}
	\hline 
	\text{Models} &\text{Style} &\text{Time(s)} & \text{Imp MSE} & \text{Bias}($\hat\beta_1$) & \text{CR($\hat\beta_1$)} & SE($\hat\beta_1$) & SD($\hat\beta_1$) \\ 
	\hline
	
	SoftImpute &SI &30.1 &0.0442 &-0.2862 &0.50 &0.1583 &0.2019 \\

    GAIN &SI &111.1 &0.7383 &0.6897 &0.18 &0.1697 &0.5693 \\
    
    MIWAE &SI &52.5 &0.1228 &0.5885 & 0.15 &0.1793 &0.2162 \\
    
    
    
    Sinkhorn &MI &39.3 &0.1031 &0.6647 &0.26 &0.2643 &0.2195 \\
    
	MI-NNGP1 &MI &4.9 &\textbf{0.0119} &\textbf{0.0351} &0.89 &0.1194 &0.1422 \\
	
	MI-NNGP1-BS &MI &4.9 &\textbf{0.0168} &\textbf{0.0383} &\textbf{0.94} &0.1424 &0.1416 \\
    
    MI-NNGP2 &MI &5.8 &\textbf{0.0086} &\textbf{0.0487} &0.90 &0.1212 &0.1343 \\
    
    MI-NNGP2-BS &MI &13.9 &\textbf{0.0092} &\textbf{0.0347} &\textbf{0.93} &0.1257 &0.1289 \\
    
	\hline 
    Complete data &-&-&-&0.0350& 0.94 &0.1122 &0.1173\\
	Complete case &-&-&-&0.2804& 0.76 &0.3466 &0.4211 \\
    ColMean Imp &SI&-&0.1130 &0.7024 &0.13 &0.2574 &  0.1957\\
	\hline 
	\end{tabular}}
\caption{Gaussian data with $n=200$ and $p=1001$ under MAR. Approximately $\textbf{40\%}$ features and $\textbf{90\%}$ cases contain missing values. Linear RR and MICE are not included due to running out of RAM. Detailed simulation setup information is in  appendix.}
\label{p1000 Gaussian MAR}
\vspace{-0.18in}
\end{table*}

\Cref{p250 Gaussian MAR} presents the results for Gaussian data with $n=200$ and $p=251$ under MAR. The MI-NNGP methods yield smallest imputation error (Imp MSE) compared to the other methods. In terms of statistical inference, the MI-NNGP methods, MICE and Sinkhorn, all of which are MI methods, lead to small to negligible bias in $\hat\beta_1$. The CR for MI-NNGP1-BS, MI-NNGP2, MI-NNGP2-BS, and Sinkhorn is close to the nominal level of $95\%$ and their SE($\hat\beta_1$) is close to SD($\hat\beta_1$), suggesting that Rubin's rule works well for these MI methods. Of these methods, our MI-NNGP methods and MICE outperform Sinkhorn in terms of information recovery, as evidenced by their smaller SE compared to Sinkhorn.  SoftImpute, Linear RR, MIWAE and GAIN, four SI methods, yield poor performance in statistical inference with considerable bias for $\hat\beta_1$ and CR away from the nominal level of $95\%$. In addition, GAIN yields substantially higher imputation error than the other methods. In terms of computation, our MI-NNGP methods are the least expensive, whereas Linear RR is the most expensive.


\Cref{p1000 Gaussian MAR} presents the results for Gaussian data with $n=200$ and $p=1001$ under MAR. As $p$ increases to 1001 from 251 in \Cref{p250 Gaussian MAR}, the performance of Sinkhorn deteriorates significantly; Linear RR and MICE run out of RAM; GAIN's performance remains poor. At the same time, our MI-NNGP methods continue to yield the most satisfactory performance. In particular, MI-NNGPs have smallest imputation error in this setting. In addition, CR($\hat{\beta}_1$) for MI-NNGP with a bootstrap step is closer to the nominal level than MI-NNGP without a bootstrap step, suggesting the bootstrap step indeed improves quantification of uncertainty of imputed values. Also, the computational time for MI-NNGP methods does not increase much as $p$ increases from 251 to 1001, demonstrating that they are scalable to ultra high-dimensional $p$--a very appealing property. This is because MI-NNGP imputes the set of features with missing values in each pattern jointly, whereas other MI methods such as MICE impute each feature iteratively.


\begin{figure*}[!htb]
\centering
  \includegraphics[width=0.28\linewidth]{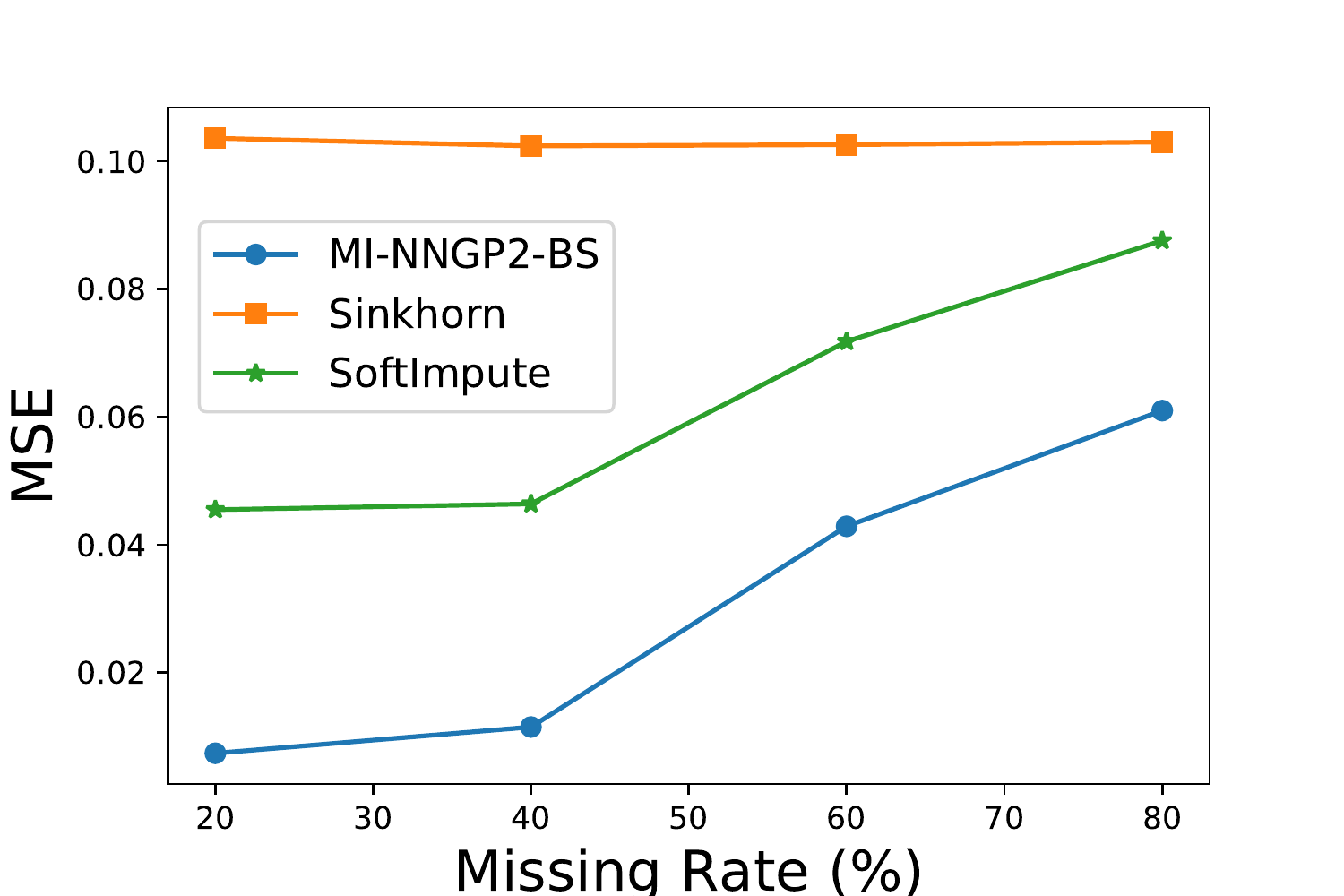}\hspace{-0.3cm}
   \includegraphics[width=0.28\linewidth]{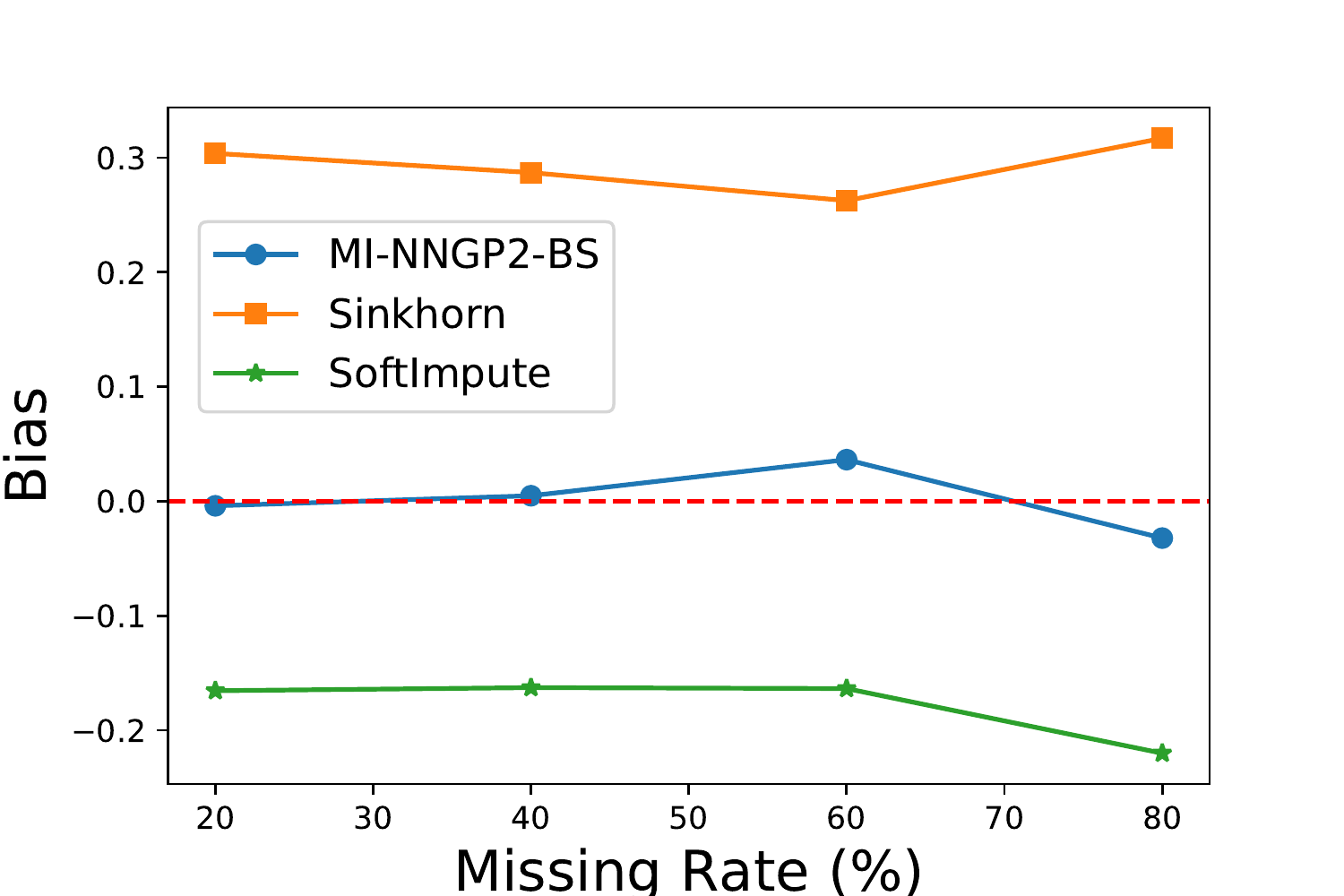}\hspace{-0.3cm}
   \includegraphics[width=0.28\linewidth]{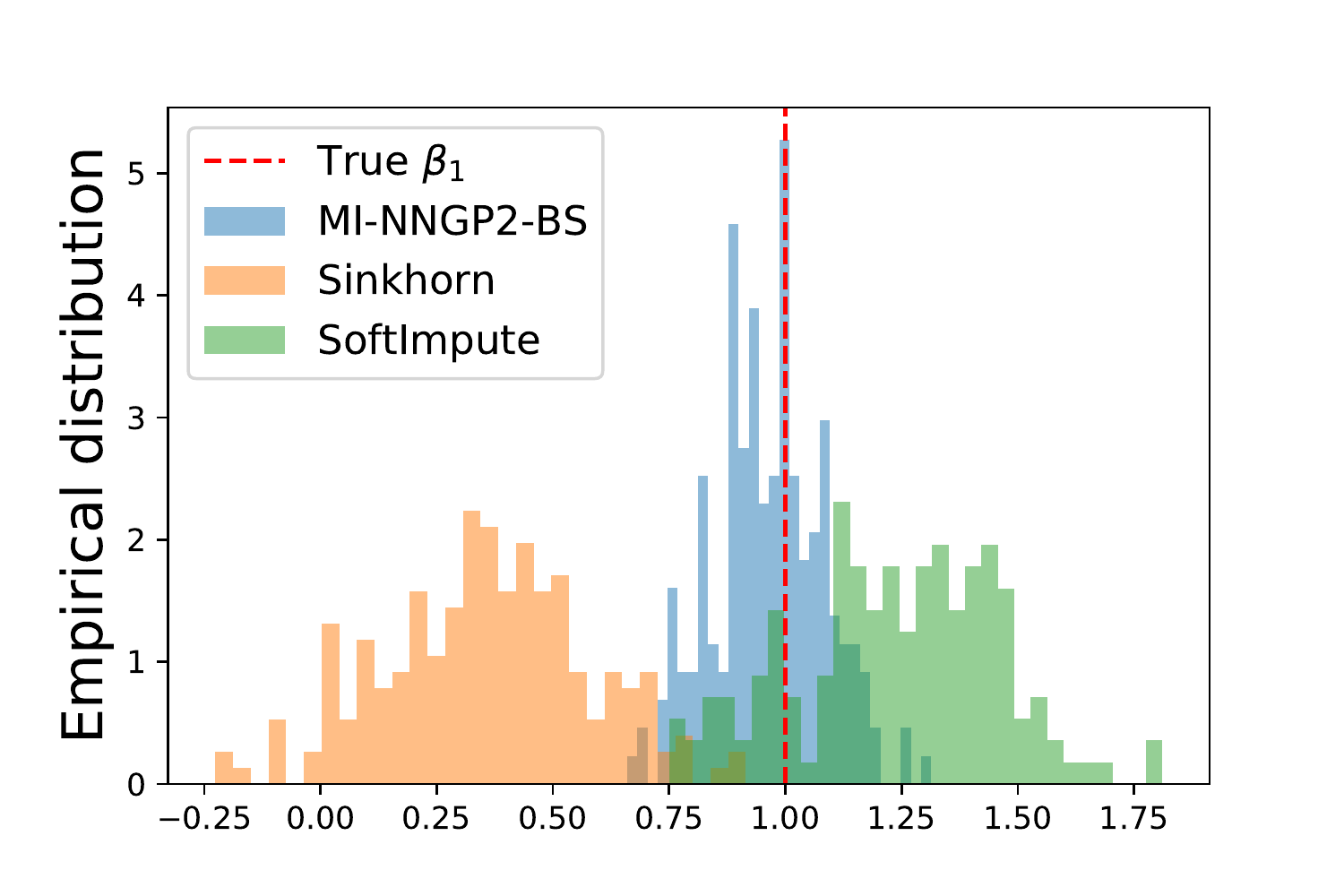}
  \caption{Left: Imputation MSE for varying missing rates. Middle: Bias of $\hat\beta_1$ for varying missing rates. Right: Empirical distribution of $\hat{\beta}_1$ from 200 MC datasets when the missing rate is $40\%$.}
\label{fig:test}
\end{figure*}

Additional results in appendix include synthetic data experiments for small $p$ under MAR (\Cref{p50 Gaussian MAR}), for \textbf{MNAR} (\Cref{p50 Gaussian MNAR}, \Cref{p250 Gaussian MNAR}, and \Cref{p1000 Gaussian MNAR}), for mix of Gaussian continuous and \textbf{discrete} data (\Cref{p1001 Gaussian MAR}, \Cref{p1001 Gaussian MNAR}), and for \textbf{non-Gaussian} continuous data (\Cref{p1000 Exponential MAR} and \Cref{p1000 exponential MNAR}). These and other unreported results for \textbf{MCAR} consistently show that the MI-NNGP methods outperform the competing state-of-the-art imputation methods, particularly in high-dimensional settings. Of the four MI-NNGP methods, MI-NNGP2-BS offers the best or close to the best performance in all experiments.

To further investigate the impact of varying missing rates on the performance of MI-NNGP2-BS, \Cref{fig:test} presents the results from additional experiments under MAR for $n=200$ and $p=1001$ in which  Sinkhorn and SoftImpute, two closest competitors based on the prior experiments, are also included. As shown in \Cref{fig:test}, MI-NNGP2-BS always yields the best performance in terms of imputation error and bias of $\hat\beta_1$ and is more robust to high missing rates.

\subsection{ADNI data}\label{sec:real missing}
\begin{table}
	\centering
	\scalebox{0.7}{
	\begin{tabular}{|c|c|c|c|c|c|c|c}
	\hline 
	\text{Models} &\text{Style} &\text{Time(s)} & \text{Imp MSE} & $\hat\beta_1$  & SE($\hat\beta_1$)  \\ 
	\hline

	SoftImpute  &SI &1008.6 &\textbf{0.0591} &0.0213 &0.0119 \\


    Sinkhorn &MI &843.3 &0.0797 & 0.0223 &0.0128 \\
   
	MI-NNGP1 &MI &7.1 &\textbf{0.0637} &\textbf{0.0161} &0.0104  \\
	
	MI-NNGP1-BS &MI &7.8 &\textbf{0.0685} &\textbf{0.0153} &0.0112  \\
	
	MI-NNGP2 &MI &11.8 &\textbf{0.0617} &\textbf{0.0171} &0.0106  \\
	
	MI-NNGP2-BS &MI &21.3 &\textbf{0.0640} &\textbf{0.0166} &0.0110  \\

	\hline 
   Complete data &-&-&-&0.0160 &0.0085  \\
	Complete case &-&-&-&0.0221 & 0.0185  \\
    ColMean Imp &SI&-&0.1534 & 0.0188 & 0.0136 \\
	\hline 
	\end{tabular}}
\caption{Real data experiment with $n=649$ and $p=10001$ under MAR. Approximately $\textbf{20\%}$ features and $\textbf{76\%}$ cases contain missing values. Linear RR, MICE, MIWAE and GAIN are not included due to running out of RAM. Detailed experiment setup information is in appendix.}
\label{adni MAR}
\vspace{-0.18in}
\end{table}

We evaluate the performance of MI-NNGPs using a publicly available, de-identified large-scale dataset from the Alzheimer's Disease Neuroimaging Initiative (ADNI), containing both image data and gene expression data.  This dataset has over 19,000 features and a response variable ($y$), VBM right hippocampal volume, for 649 patients. The details of the real data experiment are included in appendix. Briefly, we select 10000 centered features and generate missing values under MAR or MNAR. After imputation, we fit a linear regression for $y$ on three features that have highest correlation with response using imputed datasets. \Cref{adni MAR} presents the results under MAR for estimating $\beta_1$, one of the regression coefficients in the linear regression model, as well as the computational time. Again, since we do not know the true value of $\beta_1$, we cannot report its bias and instead we use $\hat\beta_1$ from the complete data analysis as a gold standard. The results in \Cref{adni MAR} show that $\hat\beta_1$ from the MI-NNGP methods is considerably closer to that from the complete data analysis than the other imputation methods, demonstrating their superior performance. In addition, SE($\hat\beta_1$ ) for MI-NNGP methods is fairly close to that for the complete data analysis and much smaller than that from the complete case analysis, suggesting that our imputation methods results in very limited information loss. In terms of computational costs, SoftImpute and Sinkhorn are much more expensive than MI-NNGP, whereas Linear RR, MICE, MIWAE and GAIN run out of memory. Additional real data experiment results in appendix under MNAR also demonstrate the superiority of our MI-NNGP methods over the existing methods.

\section{Discussion}
\label{section: discussion}
In this work, we develop powerful NNGP-based multiple imputation methods for high dimensional incomplete data with large $p$ and moderate $n$ that are also robust to high missing rates. Our experiments demonstrate that the MI-NNGP methods outperform the current state-of-the-art methods in Table 1 under MCAR, MAR and MNAR. One limitation of the MI-NNGP is that it does not scale well when $p$ becomes extremely large. To overcome this, we can take advantage of recent developments on efficient algorithms for scalable GP computation \cite{huang2015scalable,liu2020gaussian}. Instead of calculating the GP, we can approximate the GP and keep a balance between performance and computational complexity. This is our future research interest.


\bibliography{acml22}

\newpage
\appendix

\section{Details of NNGP}\label{nngp detail}
In this section, we provide the correspondence between infinitely wide fully connected neural networks and Gaussian processes which is proved in \cite{lee2017deep}. We remark that other types of neural networks, e.g. CNN, also works compatibly with the NNGP. Here we consider L-hidden-layer fully connected neural networks with input $\x \in \mathcal{R}^{d_{\text{in}}}$, layer width $n^l$ (for $l$-th layer and $d_{\text{in}}:=n^0$), parameter $\bm\theta$ consisting of weight $\W^l$ and bias $\mathbf{b}^l$ for each layer $l$ in the network, pointwise nonlinearity $\phi$, post-affine transformation (pre-activation) $z_i^l$ and post-nonlinearity $x_i^l$ for the $i$-th neuron in the $l$-th layer. We denote $x_i^0 =x_i$ for the input and use a Greek superscript $\x^{\alpha}$ to denote the $\alpha$-th sample. Weight $\W^l$ and bias $\mathbf{b}^l$ have components $W_{ij}^l$ and $b_i^l$ independently drawn from normal distribution $\mathcal{N}\left(0,\frac{\sigma_w^2}{n^l}\right)$ and $\mathcal{N}\left(0,\sigma_b^2 \right)$, respectively. 

Then the $i$-th component of pre-activation $z_i^0$ is computed as:
\begin{align*}
    z_i^0 (\x)= \sum_{j=1}^{d_{\text{in}}} W_{ij}^0 x_j +b_i^0
\end{align*}
where the pre-activation $z_i^0 (\x)$ emphasizes $z_i^0$ depends on the input $\x$. Since the weight $\W^0$ and bias $\mathbf{b}^0$ are independently drawn from normal distributions, $z_i^0 (\x)$ also follows a normal distribution. Likewise, any finite collection $\{z_i^0 (\x^{\alpha=1}),\dots,z_i^0 (\x^{\alpha=k})\}$ which is composed of $i$-th pre-activation $z_i^0$ at $k$ different inputs will have a joint multivariate normal distribution, which is exactly the definition of Gaussian process. Hence $z_i^0 \sim \mathcal{GP}(\mu^0,\mathcal{K}^0)$, where $\mu^0 (\x)= \mathbb{E}[z_i^0(\x)]=0$ and 
\begin{align*}
    \mathcal{K}^0(\x,\x') = \mathbb{E}[z_i^0(\x)z_i^0(\x')] = \sigma_b^2 + \sigma_w^2 \left(\frac{\x \cdot \x'}{d_{\text{in}}} \right)
\end{align*}
Notice that any two $z_i^0,z_j^0$ for $i \not = j$ are joint Gaussian, having zero covariance, and are guaranteed to be independent despite utilizing the same input. 

Similarly, we could analyze $i$-th component of first layer pre-activation $z_i^1$: 
\begin{align*}
    z_i^1 (\x)= \sum_{j=1}^{n^1} W_{ij}^1 x_j^1 +b_i^1 = \sum_{j=1}^{n^1} W_{ij}^1 \phi(z_j^0(\x)) +b_i^1.
\end{align*}
We obtain that $z_i^1 \sim \mathcal{GP}(0,\mathcal{K}^1)$, where
\begin{align*}
    \mathcal{K}^1(\x,\x') &=\mathbb{E}[z_i^1(\x)z_i^1(\x')] \\
    &= \sigma_b^2 + \sigma_w^2 \left(\frac{\sum_{j=1}^{n^1} \phi(z_j^0(\x))\phi(z_j^0(\x'))}{n^1} \right)
\end{align*}
Since $z_j^0 \sim \mathcal{GP}(0,\mathcal{K}^0)$, let $n^1 \to \infty$, the covariance is 
\begin{align*}
   & \mathcal{K}^1(\x,\x')\\  =& \sigma_b^2 + \sigma_w^2 \iint \phi(z)\phi(z') \\ &\mathcal{N}\left(\begin{bmatrix}
z \\
z'
\end{bmatrix};0,\begin{bmatrix}
\mathcal{K}^0(\x,\x) & \mathcal{K}^0(\x,\x') \\
\mathcal{K}^0(\x',\x) & \mathcal{K}^0(\x',\x')
\end{bmatrix} \right) \,dz \,dz'   \\
 =& \sigma_b^2 + \sigma_w^2 \mathbb{E}_{z_j^0 \sim \mathcal{GP}(0,\mathcal{K}^0)} \left[\phi(z_i^0(\x))\phi(z_i^0(\x')) \right]
\end{align*}

This integral can be solved analytically for some activation functions, such as ReLU nonlinearity \cite{cho2009kernel}. If this integral cannot be solved analytically, it can be efficiently computed numerically \cite{lee2017deep}. Hence $\mathcal{K}^1$ is determined given $\mathcal{K}^0$. 

We can extend previous arguments to general layers by induction. By taking each hidden layer width to infinity successively $(n^1\to \infty,n^2 \to \infty,\dots)$, we can conclude $z_i^l \sim \mathcal{GP}(0,\mathcal{K}^l)$, where $\mathcal{K}^l$ could be computed from the recursive relation
\begin{align*}
  \mathcal{K}^l(\x,\x')  & = \sigma_b^2 + \sigma_w^2 \mathbb{E}_{z_j^{l-1} \sim \mathcal{GP}(0,\mathcal{K}^{l-1})} \left[\phi(z_i^0(\x))\phi(z_i^0(\x')) \right]  \\
  \mathcal{K}^0(\x,\x') & = \sigma_b^2 + \sigma_w^2 \left(\frac{\x \cdot \x'}{d_{\text{in}}} \right)
\end{align*}

Hence, the covariance only depends on the neural network structure (including weight and bias variance, number of layers and activation function).

\section{Implementation details}
All the experiments run on Google Colab Pro with P100 GPU. For GAIN\footnote{See \url{https://github.com/jsyoon0823/GAIN}}, Sinkhorn, Linear RR\footnote{Same as Sinkhorn, see \url{https://github.com/BorisMuzellec/MissingDataOT}}, and MIWAE\footnote{See \url{https://github.com/pamattei/miwae}}, we use the open-access implementations provided by their authors, with the default or the recommended hyperparameters in their papers except MIWAE. For MIWAE, the default hyperparameters lead to running out RAM, hence we choose \texttt{h=128, d=10, K=20, L=1000}. For SoftImpute, the \texttt{lambda} hyperparameter is selected at each run through cross-validation and grid-point search, and we choose \texttt{maxit=500} and \texttt{thresh=1e-05}. For MICE, we use the  \texttt{iterativeImputer}\footnote{See \url{https://scikit-learn.org/stable/modules/generated/sklearn.impute.IterativeImputer.html}} method in the \texttt{scikit-learn} library with default hyperparameters \cite{pedregosa2011scikit}. All NNGP-based methods uses a 3-layer fully connected neural network with ReLU activation function to impute missing values, where the initialization of weight and bias variances are set to 1 and 0 respectively. (We also tried other initialization of weight and bias variances and found that the result is very robust to these changes.) NNGP-based methods are implemented through Neural Tangents\cite{neuraltangents2020}. All the MI methods are used to multiply impute missing values for $10$ times except GAIN, MIWAE and Linear RR, noting that the GAIN and MIWAE implementations from their authors conduct SI and Linear RR is computationally very expensive. We also include \textbf{not-MIWAE}\footnote{See \url{https://github.com/nbip/notMIWAE}} in the MNAR setting in the appendix. Similar to MIWAE, default hyperparameters of not-MIWAE lead to running out RAM, here we choose \texttt{$n_{hidden}$=128, $n_{samples}$=20, batch size=16, dl=p-1, L=1000, mprocess='selfmasking known'}. We observe that not-MIWAE is unstable and performs poorly. Probably because not-MIWAE is not scalable to high-dimensional data. 

\section{Synthetic data experiments}

\label{experiment details}

\subsection{Continuous data experiment}
\label{section: Continuous data experiment}
The simulation results are summarized over 100 Monte Carlo (MC) datasets. We also include not-MIWAE in MNAR. Note that the  Each MC dataset has a sample size of $n=200$ and includes $\y$, the fully observed outcome variable, and $\X=(\x_1,\dots,\x_p)$, the set of predictors and auxiliary variables. We consider the setting $p=50$, $p=250$, and $p=1000$ (Here the use of $p$ is a slight abuse of notation. In the main paper, $p$ represents total number features which include predictors, auxiliary variables and the response.). $\X$ is obtained by rearranging the orders of $\mathbf{A}=(\mathbf{a}_1,\dots,\mathbf{a}_p)$ and $\mathbf{A}$
is generated from a first order autoregressive model with autocorrelation $\rho$ and white noise $\epsilon$. Here $\mathbf{a}_1$ is generated from standard normal distribution $\mathcal{N}(0,1)$ if $\epsilon \sim \mathcal{N}(0,0.1^2)$ or exponential distribution $\text{Exp}(2)$ if $\epsilon \sim \text{Exp}(0.4)$. To obtain $\X$, we firstly move the fourth variable in every five consecutive variables of $\A$ (e.g. $\mathbf{a}_4$, $\mathbf{a}_9$ and $\mathbf{a}_{14}$) to the right and then the fifth variable in every five consecutive variables of $\A$ (e.g. $\mathbf{a}_5$, $\mathbf{a}_{10}$ and $\mathbf{a}_{15}$) to the right. For a concrete example, if $p=10$, $(\mathbf{a}_1,\dots,\mathbf{a}_{10})$ becomes $(\mathbf{a}_1,\mathbf{a}_{2},\mathbf{a}_{3},\mathbf{a}_{6},\mathbf{a}_{7},\mathbf{a}_{8},\mathbf{a}_{4},\mathbf{a}_{9},\mathbf{a}_{5},\mathbf{a}_{10})$ after rearrangement. The response $\y$ depends on three variables of $\X$ indicated by a set $q$: given $\X$, $\y$ is generated from
\begin{align}\label{eq: y}
    \y_i=\beta_1\cdot \x_{q[1]}+ \beta_2\cdot \x_{q[2]}  + \beta_3\cdot \x_{q[3]}  + \mathcal{N}(0,\sigma_1^2)
\end{align} 
where $\beta_i =1$ for $i\in \{1,2,3\}$. For $p=50$, $p=250$ and $p=1000$, the corresponding predictor set $q$ is $\{40,44,48\}$ $\{210,220,230\}$ and $\{650,700,750\}$ respectively. 

MAR or MNAR mechanism is considered in the simulation and the missing rate is around $40\%$. In particular, missing values are separately created in $\{\x_{\frac{3}{5}p+1},\dots,\x_{\frac{4}{5}p}\}$ and $\{\x_{\frac{4}{5}p+1},\dots,\x_p\}$ by using the following logit models for the corresponding missing indicators $\mathbf{R}_1$ and $\mathbf{R}_2$. 
If the missing mechanism is \textbf{MAR}:
\begin{align}
\label{eq: MAR indicator1}
 \text{logit}(\mathbb{P}(\mathbf{R}_1=1|\X,\y))
   &= a_1 +a_2 \cdot\frac{5}{3p}\sum_{j=1}^{3p/5}\x_{j}  +a_3\cdot\y \\
   \label{eq: MAR indicator2}
    \text{logit}(\mathbb{P}(\mathbf{R}_2=1|\X,\y))
   &=a_4 +a_5\cdot\frac{5}{3p}\sum_{j=1}^{3p/5}\x_{j}  +a_6\cdot\y
\end{align} 
If the missing mechanism is \textbf{MNAR}:
\begin{align}
\label{eq: MNAR indicator1}
 \text{logit}(\mathbb{P}(\mathbf{R}_1=1|\X,\y))
   &= a_1 +a_2*\frac{5}{p}\sum_{j=4p/5+1}^{p}\x_{j}  +a_3*\y \\
   \label{eq: MNAR indicator2}
    \text{logit}(\mathbb{P}(\mathbf{R}_2=1|\X,\y))
   &=a_4 +a_5\cdot\frac{5}{p}\sum_{j=3p/5+1}^{4p/5}\x_{j} +a_6\cdot\y
\end{align} 
If $\mathbf{R}_1=1$ or $0$, then $\{\x_{\frac{3}{5}p+1},\dots,\x_{\frac{4}{5}p}\}$ is missing or observed, respectively; similarly, if $\mathbf{R}_2=1$ or $0$, then $\{\x_{\frac{4}{5}p+1},\dots,\x_p\}$ is missing or observed, respectively. 

\subsection{Discrete data experiment}
\label{discrete data experiment}
In the discrete data analysis, we append one binary variable $\x_{p+1}$ on the last column of $\X$ in the above section. We consider the setting $p=1000$. The binary variable is generated through:
\begin{align*}
    \x_{p+1} = \begin{cases}
  1 & \text{if $\x_{10}+\x_{50}+\x_{100}>0$} \\
  0 & \text{otherwise}
\end{cases}.
\end{align*} The fully observed response $\y$ is also generated from \cref{eq: y} and the corresponding predictor set $q$ is $\{1001, 701, 751\}$. Hence $\beta_1$ is the coefficient of the binary variable in the regression model. Here missing values are separately created in $\{\x_{\frac{3}{5}p+1},\dots,\x_{\frac{4}{5}p}\}$ and $\{\x_{\frac{4}{5}p+1},\dots,\x_{p+1}\}$ with the corresponding missing indicators $\mathbf{R}_1$ and $\mathbf{R}_2$, which are also generated form \eqref{eq: MAR indicator1}, \eqref{eq: MAR indicator2} or \eqref{eq: MNAR indicator1}, \eqref{eq: MNAR indicator2} depending on the specific missing mechanism. 

Before MI-NNGPs impute, the binary variable is encoded into an one-hot, zero-mean vector (i.e. entries of -0.5 for the incorrect class and 0.5 for the correct class). After imputing this one-hot vector in the incomplete cases, the class with higher value is regarded as the imputation class.

\subsection{Experiment setting}

\begin{itemize}
    
    \item \Cref{p250 Gaussian MAR}: Continuous data experiment, MAR, $n=200$, $p=250$, $\rho=0.95$,  $\epsilon \sim \mathcal{N}(0,0.1^2)$, $\sigma_1 =0.5$, $a_1=1$, $a_2= -2$, $a_3=3$, $a_4= 0$, $a_5=2$, $a_6=-2$
    
    \item \Cref{p1000 Gaussian MAR}: Continuous data experiment, MAR, $n=200$, $p=1000$, $\rho=0.95$, $\epsilon \sim \mathcal{N}(0,0.1^2)$, $\sigma_1 =0.5$, $a_1=1$, $a_2= -2$, $a_3=3$, $a_4= 0$, $a_5=2$, $a_6=-2$
    
    \item \Cref{p1001 Gaussian MAR}: Discrete data experiment, MAR, $n=200$, $p=1000$, $\rho=0.95$, $\epsilon \sim \mathcal{N}(0,0.1^2)$, $\sigma_1 =0.5$, $a_1=-1$, $a_2= -2$, $a_3=3$, $a_4= 1$, $a_5=2$, $a_6=-2$ 
    
    \item \Cref{p50 Gaussian MAR}: Continuous data experiment, MAR, $n=200$, $p=50$, $\rho=0.95$, $\epsilon \sim \mathcal{N}(0,0.1^2)$, $\sigma_1 =0.5$, $a_1=1$, $a_2= -2$, $a_3=3$, $a_4= 0$, $a_5=2$, $a_6=-2$
        
    \item \Cref{p1000 Exponential MAR}: Continuous data experiment, MAR, $n=200$, $p=1000$, $\rho=0.75$, $\epsilon \sim \text{Exp}(0.4)$, $\sigma_1 = 1$, $a_1=-3$, $a_2= -1$, $a_3=1.5$, $a_4= 1$, $a_5=1.5$, $a_6=-1$
    
    \item \Cref{p50 Gaussian MNAR}: Continuous data experiment, MNAR, $n=200$, $p=50$, $\rho=0.95$, $\epsilon \sim \mathcal{N}(0,0.1^2)$, $\sigma_1 =0.5$, $a_1=1$, $a_2= -2$, $a_3=3$, $a_4= 0$, $a_5=2$, $a_6=-2$
    
    \item \Cref{p250 Gaussian MNAR}: Continuous data experiment, MNAR, $n=200$, $p=250$, $\rho=0.95$, $\epsilon \sim \mathcal{N}(0,0.1^2)$, $\sigma_1 =0.5$, $a_1=1$, $a_2= -2$, $a_3=3$, $a_4= 0$, $a_5=2$, $a_6=-2$
    
    \item \Cref{p1000 Gaussian MNAR}: Continuous data experiment, MNAR, $n=200$, $p=1000$, $\rho=0.95$, $\epsilon \sim \mathcal{N}(0,0.1^2)$, $\sigma_1 =0.5$, $a_1=1$, $a_2= -2$, $a_3=3$, $a_4= 0$, $a_5=2$, $a_6=-2$
    
    \item \Cref{p1000 exponential MNAR}: Continuous data experiment, MNAR, $n=200$, $p=1000$, $\rho=0.75$, $\epsilon \sim \text{Exp}(0.4)$, $\sigma_1 = 1$, $a_1=-3$, $a_2= -1$, $a_3=1.5$, $a_4= 1$, $a_5=1.5$, $a_6=-1$
    
    \item \Cref{p1001 Gaussian MNAR}: Discrete data experiment, MNAR, $n=200$, $p=1000$, $\rho=0.95$, $\epsilon \sim \mathcal{N}(0,0.1^2)$, $\sigma_1 =0.5$, $a_1=-1$, $a_2= -2$, $a_3=3$, $a_4= 1$, $a_5=2$, $a_6=-2$

\end{itemize}

    

\begin{table*}[!htb]
	\centering
	\scalebox{0.8}{
	\begin{tabular}{|c|c|c|c|c|c|c|c|c}
	\hline 
	\text{Models} &\text{Style} &\text{Time(s)} & \text{Imp MSE} & \text{Bias}($\hat\beta_1$) & \text{CR($\hat\beta_1$)} & SE($\hat\beta_1$) & SD($\hat\beta_1$) \\ 
	\hline
	
	SoftImpute &SI &2.7 &\textbf{0.0132} &\textbf{-0.0017} &\textbf{0.92} &0.0623 &0.0642 \\
	
	GAIN &SI &35.9 &1.356 & 0.3213 &  0.38 &0.1142 &0.4262 \\
	
	MIWAE &SI &46.5 &0.0361 &-0.0238 &  0.90 &0.0632 &0.0738 \\
	
	
    Linear RR &SI &628.4 &0.1712 &\textbf{0.0358} &0.91 &0.1287 &0.1568 \\
        
	MICE & MI & 2.1 & 0.0200 & \textbf{0.0031} &\textbf{0.97}  &0.0644 &0.0567 \\
    	
    Sinkhorn &MI & 42.1 &0.1081 &-0.1225 &0.60 &0.0978 &0.1269 \\

	MI-NNGP1 &MI &3.4 &\textbf{0.0129} &\textbf{0.0048} &\textbf{0.95} &0.0621 &0.0647 \\
	
	MI-NNGP1-BS &MI &4.8 &\textbf{0.0177} &\textbf{0.0052} &\textbf{0.97} &0.0794 &0.0624 \\
    
    MI-NNGP2 &MI &5.5 &\textbf{0.0092} &\textbf{0.0083} &\textbf{0.96} &0.0639 &0.0563  \\
    
    MI-NNGP2-BS &MI &13.5 &\textbf{0.0105} &\textbf{0.0083} &\textbf{0.98} &0.0705 & 0.0574  \\
    
	\hline 
    Complete data &-&-&-&0.0025 & 0.98 &0.0605 &0.0524\\
	Complete case &-&-&-&0.1869& 0.79 &0.2298 &0.2419 \\
	 ColMean Imp &SI&-&0.4716 &0.5312 &0.28 &0.2242 &  0.1729\\
	\hline 
	\end{tabular}}
\caption{Gaussian data with $n=200$ and $p=51$ under MAR. Approximately $\textbf{40\%}$ features and $\textbf{92\%}$ cases contain missing values.}
\label{p50 Gaussian MAR}
\end{table*}

\begin{table*}[!htb]
	\centering
	\scalebox{0.8}{
	\begin{tabular}{|c|c|c|c|c|c|c|c|c}
	\hline 
	\text{Models} &\text{Style} &\text{Time(s)} & \text{Imp MSE} & \text{Bias}($\hat\beta_1$) & \text{CR($\hat\beta_1$)} & SE($\hat\beta_1$) & SD($\hat\beta_1$) \\ 
	\hline
	
	SoftImpute &SI &37.8 &0.6284 &-0.5896 &0.92 &0.6348 &0.4858 \\

    GAIN &SI &130.9 & 1.691 &-0.7217 &0.40 &0.4611 &2.229 \\
    
    MIWAE &SI &58.4 &1.530 &0.5626 &  0.05 &0.1522 &0.1456 \\
    
    Sinkhorn &MI &42.1 &0.3845 &-0.2077 &1.00 &0.4566 &0.2832 \\
    
	MI-NNGP1 &MI &3.1 &0.3296 &\textbf{-0.0421} &0.75 &0.1757 &0.3220 \\
	
	MI-NNGP1-BS &MI &3.4 &0.4543 &-0.0570 &1.00 &0.3366 &0.2466 \\
    
    MI-NNGP2 &MI &3.9 &\textbf{0.2358} &0.1098 &0.80 &0.2312 &0.3383 \\
    
    MI-NNGP2-BS &MI &10.3 &\textbf{0.2516} &\textbf{-0.0242} &\textbf{0.95} &0.3203 &0.3037 \\
    
	\hline 
    Complete data &-&-&-&0.0156& 0.95 &0.0978 &0.0938\\
	Complete case &-&-&-&0.1726 & 0.89 &0.3984 &0.4534 \\
	ColMean Imp &SI&-&0.3794 &-0.1506 &1.00 &0.4556 & 0.3123\\
	\hline 
	\end{tabular}}
\caption{Exponential data with $n=200$ and $p=1001$ under MAR. Approximately $\textbf{40\%}$ features and $\textbf{92\%}$ cases contain missing values. Here Linear RR and MICE are not included due to running out of RAM.}
\label{p1000 Exponential MAR}
\end{table*}

\begin{table*}[!htb]
	\centering
	\scalebox{0.8}{
	\begin{tabular}{|c|c|c|c|c|c|c|c|c}
	\hline 
	\text{Models} &\text{Style} &\text{Time(s)} & \text{Imp MSE} & \text{Bias}($\hat\beta_1$) & \text{CR($\hat\beta_1$)} & SE($\hat\beta_1$) & SD($\hat\beta_1$) \\ 
	\hline
	
	SoftImpute &SI &2.1 &\textbf{0.0119} &\textbf{-0.0053} &\textbf{0.93} &0.0624 &0.0611 \\
	
	GAIN &SI &35.9 &1.4822 & 0.4448 & 0.24 &0.1187 &0.4641 \\
	
	MIWAE &SI &46.5 &0.0361 &-0.0238 &  0.90 &0.0632 &0.0738\\
	
	not-MIWAE &SI &40.8 &0.7566 &-0.0436 &  0.91 &0.0518 &0.0969\\
	
	Linear RR &SI &407 &0.1760 &\textbf{0.0412} &0.91 &0.1314 &0.1567 \\
    
    Sinkhorn &MI & 27.9 &0.1103 &-0.1340 &0.63 &0.1006 &0.1278 \\
    
    MICE & MI & 2.1 & 0.0198 & \textbf{0.0036} &\textbf{0.98}  &0.0636 &0.0559 \\
    
	MI-NNGP1 &MI &4.7 &\textbf{0.0130} &\textbf{0.0027} &\textbf{0.95} &0.0621 &0.0651 \\
	
	MI-NNGP1-BS &MI &3.9 &\textbf{0.0177} &\textbf{0.0026} &\textbf{0.97} &0.0799 &0.0631 \\
    
    MI-NNGP2 &MI &10.4 &\textbf{0.0088} &\textbf{0.0085} &\textbf{0.96} &0.0614 &0.0536  \\
    
    MI-NNGP2-BS &MI &10.1 &\textbf{0.0106} &\textbf{0.0093} &\textbf{0.97} &0.0711 & 0.0564  \\
    
	\hline 
    Complete data &-&-&-&0.0025 & 0.98 &0.0605 &0.0524\\
	Complete case &-&-&-&0.2143& 0.78 &0.2340 &0.4201 \\
	ColMean Imp&SI&-&0.4772 &0.5597 &0.24 &0.2246 &  0.1720\\
	\hline 
	\end{tabular}}
\caption{Gaussian data with $n=200$ and $p=51$ under MNAR. Approximately $\textbf{40\%}$ features and $\textbf{92\%}$ cases contain missing values.}
\label{p50 Gaussian MNAR}
\end{table*}

\begin{table*}[!htb]
	\centering
	\scalebox{0.8}{
	\begin{tabular}{|c|c|c|c|c|c|c|c|c}
	\hline 
	\text{Models} &\text{Style} &\text{Time(s)} & \text{Imp MSE} & \text{Bias}($\hat\beta_1$) & \text{CR($\hat\beta_1$)} & SE($\hat\beta_1$) & SD($\hat\beta_1$) \\ 
	\hline
	SoftImpute &SI &15.3 &0.0194 &-0.0997 &0.84  &0.1182 &0.1358 \\

    GAIN &SI &53.2 &0.8618 & 0.6212 & 0.18 &0.1502 &0.5088 \\
    
    MIWAE &SI &47.6 &0.0502 &0.0695 &  0.90 &0.1356 &0.1410\\
    
    not-MIWAE &SI &41.7 &1.4701 &0.1040 &  0.65 &0.1084 &0.1624\\
    
    Linear RR &SI &3009.6 &0.0658 &0.1823 &0.90 &0.1782 &0.0935 \\
    
	MICE & MI & 48.6 & 0.0233 & \textbf{-0.0049} &\textbf{0.93}  &0.1160 &0.1244 \\
    	
    Sinkhorn &MI &29.9 &0.0757 &\textbf{0.0117} &\textbf{0.97} &0.1839 &0.1523 \\
    
	MI-NNGP1 &MI &3.4 &\textbf{0.0116} &\textbf{0.0069} &\textbf{0.93} &0.1147 &0.1215 \\
	MI-NNGP1-BS &MI &3.4 &\textbf{0.0149} &\textbf{0.0140} &\textbf{0.96} &0.1285 &0.1179 \\
    
    MI-NNGP2 &MI &10.4 &\textbf{0.0085} &\textbf{-0.0024} &\textbf{0.95} &0.1123 &0.1148  \\
    
    MI-NNGP2-BS &MI &10.3 &\textbf{0.0094} &\textbf{-0.0018} &\textbf{0.96} &0.1177 &0.1147  \\
    
	\hline 
    Complete data &-&-&-&-0.0027 & 0.90 &0.1098 &0.1141\\
	Complete case &-&-&-&0.2518& 0.89 &0.3385 & 0.3319 \\
	ColMean Imp&SI&-& 0.1414 & 0.3539 &0.72 &0.2210 &  0.1712\\
	\hline 
	\end{tabular}}
\caption{Gaussian data with $n=200$ and $p=251$ under MNAR. Approximately $\textbf{40\%}$ features and $\textbf{90\%}$ cases contain missing values.}
\label{p250 Gaussian MNAR}
\end{table*}

\begin{table*}[!htb]
	\centering
	\scalebox{0.8}{
	\begin{tabular}{|c|c|c|c|c|c|c|c|c}
	\hline 
	\text{Models} &\text{Style} &\text{Time(s)} & \text{Imp MSE} & \text{Bias}($\hat\beta_1$) & \text{CR($\hat\beta_1$)} & SE($\hat\beta_1$) & SD($\hat\beta_1$) \\ 
	\hline
	
	SoftImpute &SI &25.1 &0.0443 &-0.2550 &0.52 &0.1570 &0.2164 \\

    GAIN &SI &111.1 &0.7395 &0.6488 &0.18 &0.1719 &0.5830 \\
    
    MIWAE &SI &53.2 &0.1116 &0.5249 &  0.35 &0.1902 &0.2523 \\
    
    not-MIWAE &SI &45.4 &3.981 &0.9897 &  0.0 &0.1080 &0.1451 \\
    
    Sinkhorn &MI &116.9 &0.0889 &0.5445 &0.38 &0.2406 &0.2237 \\
    
	MI-NNGP1 &MI &4.9 &\textbf{0.0119} &\textbf{0.0351} &0.89 &0.1194 &0.1422 \\
	
	MI-NNGP1-BS &MI &4.9 &\textbf{0.0166} &\textbf{0.0383} &\textbf{0.94} &0.1424 &0.1416 \\
    
    MI-NNGP2 &MI &10.5 &\textbf{0.0085} &\textbf{0.0356} &0.91 &0.1160 &0.1310 \\
    
    MI-NNGP2-BS &MI &9.9 &\textbf{0.0092} &\textbf{0.0343} &\textbf{0.93} &0.1263 &0.1301 \\
    
	\hline 
    Complete data &-&-&-&0.0350& 0.94 &0.1122 &0.1173\\
	Complete case &-&-&-&0.2824& 0.76 &0.3447 &0.4201 \\
	ColMean Imp&SI&-&0.1130 &0.7022 &0.11 &0.2572 &  0.1941\\
	\hline 
	\end{tabular}}
\caption{Gaussian data with $n=200$ and $p=1001$ under MNAR. Approximately $\textbf{40\%}$ features and $\textbf{90\%}$ cases contain missing values. Here Linear RR and MICE are not included due to running out of RAM.}
\label{p1000 Gaussian MNAR}
\end{table*}

	

    
    
	
    
    
    

\begin{table*}[!htb]
	\centering
	\scalebox{0.8}{
	\begin{tabular}{|c|c|c|c|c|c|c|c|c}
	\hline 
	\text{Models} &\text{Style} &\text{Time(s)} & \text{Imp MSE} &\text{Bias}($\hat\beta_1$) & \text{CR($\hat\beta_1$)} & SE($\hat\beta_1$) & SD($\hat\beta_1$) \\ 
	\hline
	
	SoftImpute &SI & 39.1 &0.6682 &-0.6784 &0.90 &0.6805 &0.4632 \\

    GAIN &SI &91.0 & 1.6974 &0.0331 &0.33 &0.4187 &2.2123 \\
    
   MIWAE &SI &57.4 &1.4937 &0.3981 &  0.30 &0.1437 &0.1659 \\
    
    not-MIWAE &SI &43.6 &26.7277 &0.7388 &  0.00 &0.0928 &0.1524 \\
    
    Sinkhorn &MI &53.2 &0.3837 &-0.2698 &1.0 &0.4362 &0.2886 \\
    
	MI-NNGP1 &MI &3.5 &0.3296 &\textbf{-0.0354} &0.76 &0.1764 &0.3166 \\
	
	MI-NNGP1-BS &MI &3.6 &0.4545 &-0.0546 &0.99 &0.3324 &0.2466 \\
    
    MI-NNGP2 &MI &4.1 &\textbf{0.2360} &0.0835 &0.82 &0.2301 &0.3327 \\
    
    MI-NNGP2-BS &MI &10.3 &\textbf{0.2501} &\textbf{-0.0449} &\textbf{0.94} &0.3183 &0.3043 \\
    
	\hline 
    Complete data &-&-&-&0.0156& 0.95 &0.0978 &0.0938\\
	Complete case &-&-&-&0.1663& 0.89 &0.4038 &0.4639 \\
	ColMean Imp &SI&-&0.3793 &-0.1574 &1.0 &0.4559 &  0.3087\\
	\hline 
	\end{tabular}}
\caption{Exponential data with $n=200$ and $p=1001$ under MNAR. Approximately $\textbf{40\%}$ features and $\textbf{92\%}$ cases contain missing values. Here Linear RR and MICE are not included due to running out of RAM.}
\label{p1000 exponential MNAR}
\end{table*}

\begin{table*}[!htb]
	\centering
	\scalebox{0.8}{
	\begin{tabular}{|c|c|c|c|c|c|c|c|c|c|}
	\hline 
	\text{Models} &\text{Style} &\text{Time(s)} & \text{Imp MSE} & Imp accu & \text{Bias}($\hat\beta_1$) & \text{CR($\hat\beta_1$)} & SE($\hat\beta_1$) & SD($\hat\beta_1$) \\ 
	\hline
	
	SoftImpute &SI &21.8 &0.0431 &0.3331 &-0.2793 &0.64  &0.2156 &0.2263 \\

    GAIN &SI &98.9 &0.8942 &0.3331 &0.8610 &0.24 &0.1588 &0.7651 \\
    
    MIWAE &SI &57.7&0.1267 &0.6785 &0.6658 &0.04 &0.1617 &0.2030 \\
    
    
    Sinkhorn &MI &41.6 &0.1076 &0.3278 &0.4201 &0.73 &0.3036 &0.2883 \\
    
	MI-NNGP1 &MI &4.7 &\textbf{0.0116} &0.6463 &\textbf{-0.0147} &\textbf{0.94} &0.1492 &0.1495 \\
	
	MI-NNGP1-BS &MI &3.4 &\textbf{0.0145} &0.6188  &\textbf{-0.0115} &\textbf{0.98} &0.1771 &0.1436 \\
    
    MI-NNGP2 &MI &3.9 &\textbf{0.0089} &\textbf{0.7289} &\textbf{0.0126} &0.99 &0.1470 &0.1247 \\
    
    MI-NNGP2-BS &MI &9.5 &\textbf{0.0093} &\textbf{0.7006} &\textbf{-0.0014} &\textbf{0.98}  &0.1556 &0.1258 \\
    
	\hline 
    Complete data &-&-&-&-&0.0156& 0.96 &0.1119 &0.1041\\
	Complete case &-&-&-&-&0.3856 & 0.70 &0.2846 &0.2937 \\
	ColMean Imp &SI&-&0.1255&0.3278 &0.4643 &0.71 &0.3000 & 0.2362\\
	\hline 
	\end{tabular}}
\caption{Gaussian and binary data with $n=200$ and $p=1002$ under MAR. Approximately $\textbf{40\%}$ features and $\textbf{88\%}$ cases contain missing values.  Linear RR and MICE are not included due to running out of RAM. Detailed simulation setup information is in appendix. }
\label{p1001 Gaussian MAR}
\end{table*}

\begin{table*}[!htb]
	\centering
	\scalebox{0.8}{
	\begin{tabular}{|c|c|c|c|c|c|c|c|c|c|}
	\hline 
	\text{Models} &\text{Style} &\text{Time(s)} & \text{Imp MSE}& Imp accu & \text{Bias}($\hat\beta_1$) & \text{CR($\hat\beta_1$)} & SE($\hat\beta_1$) & SD($\hat\beta_1$) \\ 
	\hline
	
	SoftImpute &SI &24.7 &0.0432 &0.3328 &-0.2771 &0.64 &0.2154 &0.2263 \\

    GAIN &SI &95.3 &0.8517 &0.3328 &0.9779 &0.12 &0.1605 &0.7701 \\
    
    MIWAE &SI &57.7&0.1258 &0.6706 &0.6665 &0.15 &0.1604 &0.2274 \\
    
   not-MIWAE &SI &42.7 &1.9305 &0.3317 &0.6158 &0.06 &0.0851 &0.1876 \\
    
    Sinkhorn &MI &43.6 &0.1080 &0.3273 &0.4175 &0.72  &0.3033 &0.2882 \\
    
	MI-NNGP1 &MI &3.6 &\textbf{0.0117} &0.6477 &\textbf{-0.0137} &\textbf{0.94} &0.1492 &0.1514 \\
	
	MI-NNGP1-BS &MI &3.7 &\textbf{0.0146} &0.6201 &\textbf{-0.0154} &\textbf{0.98}  &0.1785 &0.1423 \\
    
    MI-NNGP2 &MI &4.2 &\textbf{0.0089} &\textbf{0.7277} &\textbf{0.0125} &0.99  &0.1469 &0.1238 \\
    
    MI-NNGP2-BS &MI &10.0 &\textbf{0.0093} &\textbf{0.6971} &\textbf{-0.0044} &\textbf{0.98} &0.1558 &0.1243 \\
    
	\hline 
    Complete data &-&-&-&-&0.0156& 0.96 &0.1119 &0.1041\\
	Complete case &-&-&-&0.3909 & 0.70 &- &0.2854 &0.3001 \\
	ColMean Imp &SI&-&0.1259 &0.3273 &0.4620 &0.71  &0.2996 & 0.2367 \\
	\hline 
	\end{tabular}}
\caption{Gaussian and binary data with $n=200$ and $p=1002$ under MNAR. Approximately $\textbf{40\%}$ features and $\textbf{88\%}$ cases contain missing values. Here Linear RR and MICE are not included due to running out of RAM.}
\label{p1001 Gaussian MNAR}
\end{table*}

\subsection{Varying missing rates experiment}
\label{section: varying missing rates}
Here we state clearly the varying missing rates experiment. Similar to the data generation process in the continuous data experiment, each MC dataset has sample size of $n=200$ and each sample includes a response $y$ and $p=1000$ features. When generating variable set $\A$, $\mathbf{a}_1$ is drawn from $\mathcal{N}(0,1)$ and the remaining variables are generated through first order autoregressive model with autocorrelation $\rho=0.95$ and white noise $\mathcal{N}(0,0.1^2)$. $\X$ is obtained by firstly moving the seventh variable and ninth variable in every ten consecutive variables of $\A$ (e.g., $\mathbf{a}_7$, $\mathbf{a}_9$, $\mathbf{a}_{17}$ and $\mathbf{a}_{19}$) to the right and then the eighth variable and tenth variable in every ten consecutive variables of $\A$ (e.g., $\mathbf{a}_8$, $\mathbf{a}_{10}$, $\mathbf{a}_{18}$ and $\mathbf{a}_{20}$) to the right. Given $\X$, $\y$ is generated from \eqref{eq: y} with corresponding predictor set $q=\{910,950,990\}$. Missing values are separately created in two groups of variables under MAR by using the following logit models for the corresponding missing indicators $\mathbf{R}_1$ and $\mathbf{R}_2$:
\begin{align*}
 \text{logit}(\mathbb{P}(\mathbf{R}_1=1|\X,\y))
   &= 1 -\frac{1}{50}\sum_{j=1}^{100}\x_{j}  +3\y \\
    \text{logit}(\mathbb{P}(\mathbf{R}_2=1|\X,\y))
   &=\frac{1}{50}\sum_{j=1}^{100}\x_{j} -2\y
\end{align*} 
If the missing rate is $20\%$, the first group is $\{\x_{801},\dots,\x_{900}\}$ and the second group is $\{\x_{901},\dots,\x_{1000}\}$. If the missing rate is $40\%$, the first group is $\{\x_{601},\dots,\x_{800}\}$ and the second group is $\{\x_{801},\dots,\x_{1000}\}$. If the missing rate is $60\%$, the first group is $\{\x_{401},\dots,\x_{700}\}$ and the second group is $\{\x_{701},\dots,\x_{1000}\}$. If the missing rate is $80\%$, the first group is $\{\x_{201},\dots,\x_{600}\}$ and the second group is $\{\x_{601},\dots,\x_{1000}\}$.

\section{ADNI data experiments}
\label{adni experiment detail}

\subsection{Data Availability}
The de-identified ADNI dataset is publicly available at \url{http://adni.loni.usc.edu/}. 

\subsection{Experiment details}
This section details the ADNI data experiment. Here we use a large-scale dataset from ADNI study. The original dataset includes 19822 features and one continuous response variable ($\y$), the VBM right hippocampal volume, for 649 patients. Here we preprocess features and the response by removing their means. Among these 19822 features, we only select 10000 features which have maximal correlation with the response to analyze and rank them in the decreasing order of correlation. Denote the selected features by $\X=(\x_1,\dots,\x_{10000})$. In the analysis model, the first three features are chosen as predictors and our goal is to fit the regression model $\mathbb{E}[\y|\x_1,\x_2,\x_3] = \beta_0 + \beta_1 \x_1+\beta_2\x_2+\beta_3\x_3$ and analyze the first coefficient $\beta_1$.  

There are no missing values in the original data, so we artificially introduce some missing values, which are separately created in two groups: $\{\x_1,\dots,\x_{1000}\}$ and $\{\x_{1001},\dots,\x_{2000}\}$ by the following logit models for the corresponding missing indicator $\mathbf{R}_1$ and $\mathbf{R}_2$. If the missing mechanism is \textbf{MAR}:
\begin{align*}
    \text{logit}(\mathbb{P}(\mathbf{R}_1=1)) &=-1-\frac{3}{100}\sum_{j=2001}^{2100}\x_{j}+3\y \\
   \text{logit}(\mathbb{P}(\mathbf{R}_2=1)) &=-1-\frac{3}{100}\sum_{j=2201}^{2300}\x_{j}+2\y 
\end{align*}
If the missing mechanism is \textbf{MNAR}:
\begin{align*}
    \text{logit}(\mathbb{P}(\mathbf{R}_1=1)) &=-1-\frac{3}{5}\sum_{j=1001}^{1005}\x_{j}+3\y \\
   \text{logit}(\mathbb{P}(\mathbf{R}_2=1)]) &=-1-\frac{3}{5}\sum_{j=1}^{5}\x_{j}+2\y 
\end{align*}

We repeat the above procedure for 100 times to generate 100 incomplete datasets. Each incomplete dataset only differs in location of missing values and therefore they are not Monte Carlo datasets (which is the reason that we do not include the SD($\hat\beta_1$) in this experiment). We impute the incomplete datasets and present the summarized results. 
\begin{table}[!htb]
	\centering
	\scalebox{0.8}{
	\begin{tabular}{|c|c|c|c|c|c|c|c}
	\hline 
	\text{Models} &\text{Style} &\text{Time(s)} & \text{Imp MSE} & $\hat\beta_1$  & SE($\hat\beta_1$)  \\ 
	\hline

	SoftImpute  &SI &991.5 &\textbf{0.0613} &0.0212 &0.0114 \\

    Sinkhorn &MI &709.8 &0.0866 & 0.0216 &0.0123  \\
   
	MI-NNGP1 &MI &7.4 &\textbf{0.0644} &\textbf{0.0155} &0.0101  \\
	
	MI-NNGP1-BS &MI &7.7 &\textbf{0.0688} &\textbf{0.0162} &0.0112  \\
	
	MI-NNGP2 &MI &11.6 &\textbf{0.0622} &\textbf{0.0145} &0.0103  \\
	
	MI-NNGP2-BS &MI &18.5 &\textbf{0.0609} &0.0123 &0.0125  \\

	\hline 
   Complete data&-&-&-&0.0160 &0.0085  \\
	Complete case &-&-&-&0.0202 & 0.0172  \\
	ColMean Imp&SI&-&0.1685 & 0.01776 & 0.0130 \\
	\hline 
	\end{tabular}}
\caption{Real data experiment with $n=649$ and $p=10001$ under MNAR. Approximately $\textbf{20\%}$ features and $\textbf{74\%}$ cases contain missing values. Linear RR, MICE, not-MIWAE and GAIN are not included due to running out of RAM.}
\label{ADNI MNAR}
\end{table}

\end{document}